\documentclass[smallextended]{svjour3}       
\smartqed  
\usepackage{graphicx}
\usepackage{times}
\usepackage{helvet}
\usepackage{courier}
\usepackage{algorithmic}
\usepackage{algorithm}
\usepackage{epsfig}
\usepackage{amssymb}
\usepackage{amsmath}
\usepackage{stfloats}
\usepackage{multirow}
\usepackage{tikz}

\def\dup{\tt DUP}
\def\rnnplanner{\tt RNNPlanner}
\def\caseA{\textbf{\textit{case A}}}
\def\caseB{\textbf{\textit{case B}}}
\def\caseC{\textbf{\textit{case C}}}
\def\caseD{\textbf{\textit{case D}}}
\def\lib{\mathcal{L}}
\def\obs{\mathcal{O}}
\def\a{\mathcal{A}}
\def\O{\phi}
\def\uplan{\tilde{p}}
\def\one{\mathbb{I}}
\def\weight{\Gamma}

\pdfoutput=1

\newcommand{\ignore}[1]{{}}

\begin{document}

\title{Discovering Underlying Plans Based on Shallow Models}

\author{Hankz Hankui Zhuo         \and
        Yantian Zha  \and 
        Subbarao Kambhampati
}

\institute{ Hankz Hankui Zhuo \at
              Sun Yat-Sen University \\
              \email{zhuohank@mail.sysu.edu.cn}  
              \and
              Yantian Zha \at
              Arizona State University \\
              \email{yantian.zha@asu.edu}
              \and
              Subbarao Kambhampati \at
              Arizona State University \\
              \email{rao@asu.edu}
}


\maketitle

\begin{abstract}
Plan recognition aims to discover target plans (i.e., sequences of actions) behind observed actions, with history plan libraries or domain models in hand. Previous approaches either discover plans by maximally ``matching'' observed actions to plan libraries, assuming target plans are from plan libraries, or infer plans by executing domain models to best explain the observed actions, assuming that complete domain models are available. In real world applications, however, target plans are often not from plan libraries, and complete domain models are often not available, since building complete sets of plans and complete domain models are often difficult or expensive. In this paper we view plan libraries as corpora and learn vector representations of actions using the corpora; we then discover target plans based on the vector representations. Specifically, we propose two approaches, {\dup} and {\rnnplanner}, to discover target plans based on vector representations of actions. {\dup} explores the EM-style framework to capture local contexts of actions and discover target plans by optimizing the probability of target plans, while {\rnnplanner} aims to leverage long-short term contexts of actions based on RNNs (recurrent neural networks) framework to help recognize target plans. In the experiments, we empirically show that our approaches are capable of discovering underlying plans that are not from plan libraries, without requiring domain models provided. We demonstrate the effectiveness of our approaches by comparing its performance to traditional plan recognition approaches in three planning domains. We also compare {\dup} and {\rnnplanner} to see their advantages and disadvantages. 

\end{abstract}

\section{Introduction}
As computer-aided cooperative work scenarios become increasingly popular, human-in-the-loop planning and decision support has become a critical planning challenge (c.f. \cite{cacm-sketch-plan,woogle,ai-mix}). An important aspect of such a support \cite{aaai-hilp-tutorial} is recognizing what plans the human in the loop is making, and provide appropriate suggestions about their next actions \cite{DBLP:conf/uai/AlbrechtR15}.
Although there is a lot of work on plan recognition, much of it has
traditionally depended on the availability of a complete domain model
\cite{geffner-ramirez,conf/nips/hankz12,DBLP:journals/ai/ZhuoK17}. As has been argued elsewhere
\cite{aaai-hilp-tutorial}, such models are hard to get
in human-in-the-loop planning scenarios. Here, the decision support
systems have to make themselves useful without insisting on complete
action models of the domain. The situation here is akin to that faced
by search engines and other tools for computer supported
cooperate work, and is thus a significant departure for the ``planning
as pure inference'' mindset of the automated planning community. As
such, the problem 
calls for plan recognition with ``shallow'' models of the domain
(c.f. \cite{rao-model-lite}), that can be easily learned
automatically. Compared to learning action models (``complex'' models correspondingly) of the domain from limited training data, learning shallow models can avoid the overfitting issue. One key difference between ``shallow'' and ``complex'' models is the size of parameters of both models is distinguish, which is comparable to learning models in machine learning community, i.e., complex models with large parameters require much more training data for learning parameter values compared to ``shallow'' models.

\begin{figure}[!ht]
\centerline{\includegraphics[width=0.5\textwidth]{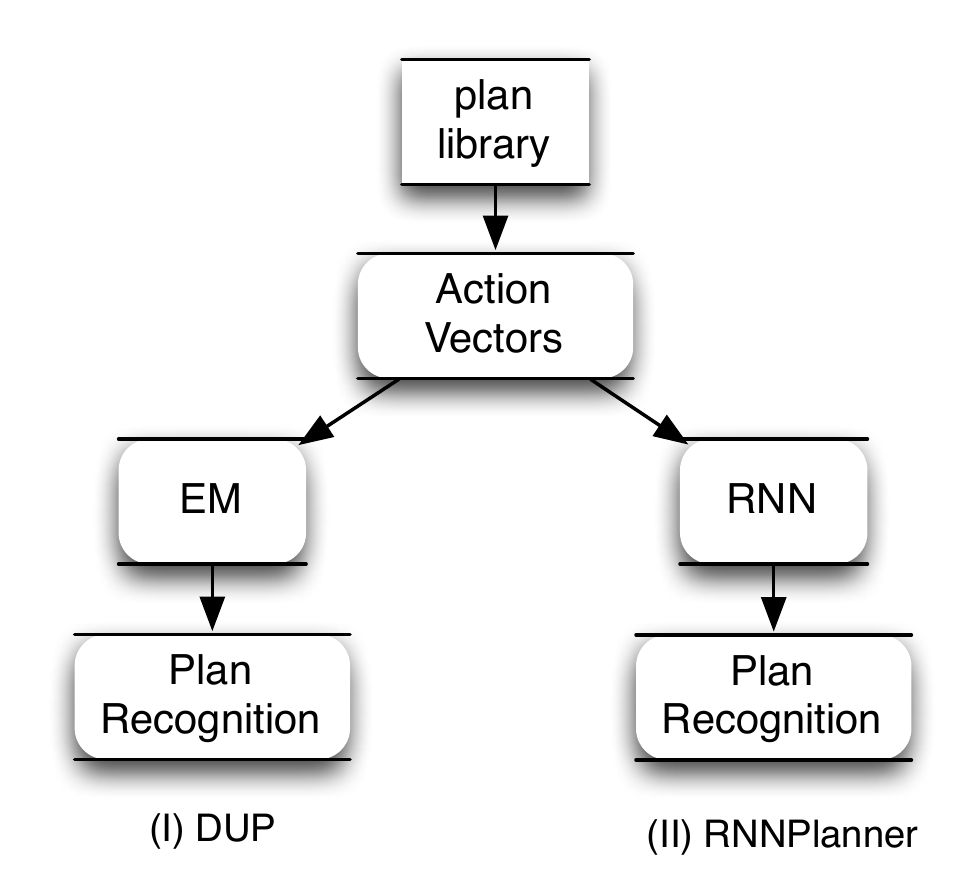}}
\caption{The framework of our shallow models {\dup} and {\rnnplanner}}
\label{framework}
\end{figure}

There has been very little work on learning such shallow models to support human-in-the-loop planning. Some examples include the work on Woogle system \cite{woogle} that aimed to provide support to humans in web-service composition. That work however relied on very primitive understanding of the actions (web services in their case) that consisted merely of learning the input/output types of individual services. In this paper, we focus on learning more informative models that can help recognize the plans under construction by the humans, and provide active support by suggesting relevant actions. To drive this process, we propose two approaches to learning informative models, namely {\dup}, standing for \textbf{D}iscovering \textbf{U}nderlying \textbf{P}lans based on action-vector representations, and {\rnnplanner}, standing for \textbf{R}ecurrent \textbf{N}eural \textbf{N}etwork based \textbf{Planner}. The framework of {\dup} and {\rnnplanner} is shown in Figure \ref{framework}, where we take as input a set of plans (or a \emph{plan library}) and learn the distributed representations of actions (namely \emph{action vectors}). After that, our {\dup} approach exploits an EM-Style framework to discover underlying plans based on the learnt action vectors, while our {\rnnplanner} approach exploits an RNN-Style framework to generate plans to best explain observations (i.e., discover underlying plans behind the observed actions) based on the learnt action vectors. In {\dup} we consider local contexts (with a limited window size) of actions being recognized, while in {\rnnplanner} we explore the potential influence from long and short-term actions, which can be modelled by RNN, to help recognize unknown actions. 

In summary, the contributions of the paper are shown below.
\begin{enumerate}
\item In \cite{DBLP:conf/atal/TianZK16}, we presented a version of {\dup}. In this paper we extend \cite{DBLP:conf/atal/TianZK16} with more details to elaborate the approach.
\item We propose a novel model {\rnnplanner} based on RNN to explore the influence of actions from long and short-term contexts. 
\item We compare {\rnnplanner} to {\dup} to exhibit the advantage and disadvantage of leveraging information from long and short-term contexts. 
\end{enumerate}

In the sequel, we first formulate our plan recognition problem, and then address the details of our approaches {\dup} and {\rnnplanner}. After that, we empirically demonstrate that it does capture a surprising amount of structure in the observed plan sequences, leading to effective plan recognition. We further compare its performance to traditional plan recognition techniques, including one that uses the same plan traces to learn the STRIPS-style action models, and use the learned model to support plan recognition. We also compare {\rnnplanner} with {\dup} to see the advantage and disadvantage of leveraging long and short-term contexts of actions in different scenarios. We finally review previous approaches related to our work and conclude our paper with further work.


\section{Problem Formulation}
A plan library, denoted by $\lib$, is composed of a set of plans $\{p\}$, where $p$ is a sequence of actions, i.e., $p=\langle a_1,a_2,\ldots,a_n\rangle$ where $a_i$, $1\leq i\leq n$, is an action name (without any parameter) represented by a string. For example, a string \emph{unstack-A-B} is an action meaning that \emph{a robot unstacks block A from block B}. We denote the set of all possible actions by $\bar{\a}$ which is assumed to be known beforehand. For ease of presentation, we assume that there is an empty action, $\O$, indicating an unknown or not observed action, i.e., $\a=\bar{\a}\cup\{\O\}$. An observation of an \emph{unknown} plan $\uplan$ is denoted by $\obs=\langle o_1,o_2,\ldots,o_M\rangle$, where $o_i\in\a$, $1\leq i\leq M$, is either an action in $\bar{\a}$ or an empty action $\O$ indicating the corresponding action is missing or not observed. Note that $\uplan$ is not necessarily in the plan library $\lib$, which makes the plan recognition problem more challenging, since matching the observation to the plan library will not work any more.

We assume that the human is making a plan of at most length $M$. We also
assume that at any given point, the planner is able to observe $M-k$ of
these actions. The $k$ unobserved actions might either be in the suffiix of the plan, or in the middle.
Our aim is to suggest, for each of the $k$ unobserved actions, $m$
possible choices from which the user can select the action. (Note
that we would like to keep $m$ small, ideally close to 1, so as not to
overwhelm users).
Accordingly, we will evaluate the effectiveness of the decision
support in terms of whether or not the user's best/intended action is
within the suggested $m$ actions.

Specifically, our recognition problem can be represented by a triple $\Re=(\lib,\obs,\a)$. The solution to $\Re$ is to discover the unknown plan $\uplan$, which is a plan with unkwown observations, that best explains $\obs$ given $\lib$ and $\a$. We have the following assumptions \textbf{A1-A3}:
\begin{itemize}
\item[A1:] The length of the underlying plan to be discovered is known, which releases us from searching unlimited length of plans.
\item[A2:] The positions of missing actions in the underlying plan is known in advance, which releases us from searching missing actions in between observed actions.
\item[A3:] All actions observed are assumed to be correct, which indicates there is no need to criticize or rectify the observed actions.
\end{itemize}
An example of our plan recognition problem in the \emph{blocks}\footnote{http://www.cs.toronto.edu/aips2000/} domain is shown below. 

\textbf{Example:} A plan library $\lib$ in the \emph{blocks} domain is assumed to have four plans as shown below:
\begin{center}
\begin{tabular}{|l|}
\hline
\textbf{plan 1}: \emph{pick-up-B stack-B-A pick-up-D stack-D-C} \\
\textbf{plan 2}: \emph{unstack-B-A put-down-B unstack-D-C put-down-D} \\
\textbf{plan 3}: \emph{pick-up-B stack-B-A pick-up-C stack-C-B pick-up-D stack-D-C} \\
\textbf{plan 4}: \emph{unstack-D-C put-down-D unstack-C-B put-down-C unstack-B-A put-down-B} \\
\hline
\end{tabular}
\end{center}
An observation $\obs$ of action sequence is shown below:
\begin{center}
\begin{tabular}{|l|}
\hline
\textbf{observation:} \emph{pick-up-B  $\O$ unstack-D-C put-down-D  $\O$ stack-C-B $\O$ $\O$} \\
\hline
\end{tabular}
\end{center}
Given the above input, our {\dup} algorithm outputs plans as follows:
\begin{center}
\begin{tabular}{|p{0.85\textwidth}|}
\hline
\emph{pick-up-B stack-B-A unstack-D-C put-down-D pick-up-C stack-C-B pick-up-D stack-D-C}  \\
\hline
\end{tabular}
\end{center}

Although the ``plan completion'' problem seems to differ superficially
from the traditional ``plan recognition'' problem, we point out that
many earlier works on plan recognition do in fact evaluate their
recognition algorithms in terms of completion tasks, e.g.,
\cite{cof/ijcai/Ramirez09,cof/ijcai/hankz11,conf/nips/hankz12}. While
these earlier efforts use different problem settings, taking either a
plan library or action models as input, they share one common
characteristic: they all aim to look for a plan that can best explain
(or complete) the observed actions. This is exactly the same as our
problem we aim to solve.

\section{Learning the distributed representations of actions} \label{actionvector}
Since actions are denoted by a name string, actions can be viewed as words, and a plan can be viewed as a sentence. Furthermore, the plan library $\lib$ can be seen as a corpus, and the set of all possible actions $\a$ is the vocabulary. Given a plan corpus, we can exploit off-the-shelf approaches, e.g., the Skip-gram model \cite{word2vec}, for learning vector representations for actions. 

The objective of the Skip-gram model is to learn vector representations for predicting the surrounding words in a sentence or document. Given a corpus $\mathcal{C}$, composed of a sequence of training words $\langle w_1,w_2,\ldots,w_T\rangle$, where $T=|\mathcal{C}|$, the Skip-gram model maximizes the average log probability 
\begin{equation}\label{skip-gram}
\frac{1}{T}\sum_{t=1}^T\sum_{-c\leq j\leq c,j\neq0}\log p(w_{t+j}|w_t)
\end{equation} 
where $c$ is the size of the training window or context. 

The basic probability $p(w_{t+j}|w_t)$ is defined by the hierarchical softmax, which uses a binary tree representation of the output layer with the $K$ words as its leaves and for each node, explicitly represents the relative probabilities of its child nodes \cite{word2vec}. For each leaf node, there is an unique path from the root to the node, and this path is used to estimate the probability of the word represented by the leaf node. There are no explicit output vector representations for words. Instead, each inner node has an output vector $v'_{n(w,j)}$, and the probability of a word being the output word is defined by
\begin{eqnarray}\label{prediction}
p(w_{t+j}|w_t)=\prod_{i=1}^{L(w_{t+j})-1}\Big\{\sigma(\one(n(w_{t+j},i+1)= \notag\\ child(n(w_{t+j},i)))\cdot v_{n(w_{t+j},i)}\cdot v_{w_t})\Big\},
\end{eqnarray}
where \[\sigma(x)=1/(1+\exp(-x)).\] $L(w)$ is the length from the root to the word $w$ in the binary tree, e.g., $L(w)=4$ if there are four nodes from the root to $w$. $n(w,i)$ is the $i$th node from the root to $w$, e.g., $n(w,1)=root$ and $n(w,L(w))=w$. $child(n)$ is a fixed child (e.g., left child) of node $n$.  $v_{n}$ is the vector representation of the inner node $n$. $v_{w_t}$ is the input vector representation of word $w_t$. The identity function $\one(x)$ is 1 if $x$ is true; otherwise it is -1.

We can thus build vector representations of actions by maximizing Equation (\ref{skip-gram}) with corpora or plan libraries $\lib$ as input. We will exploit the vector representations to discover the unknown plan $\uplan$ in the next subsection.


\section{Our {\dup} Algorithm}

Our {\dup} approach to the recognition problem $\Re$ functions by two phases. We first learn vector representations of actions using the plan library $\lib$. We then iteratively sample actions for unobserved actions $o_i$ by maximizing the probability of the unknown plan $\uplan$ via the EM framework. We present {\dup} in detail in the following subsections.


\subsection{Maximizing Probability of Unknown Plans}
With the vector representations learnt in the last subsection, a straightforward way to discover the unknown plan $\uplan$ is to explore all possible actions in $\bar{\a}$ such that $\uplan$ has the highest probability, which can be defined similar to Equation (\ref{skip-gram}), i.e.,  
\begin{equation}\label{basic-way}
\mathcal{F}(\uplan)=\sum_{k=1}^M\sum_{-c\leq j\leq c,j\neq0} \log p(w_{k+j}|w_k)
\end{equation}
where $w_k$ denotes the $k$th action of $\uplan$ and $M$ is the length of $\uplan$. As we can see, this approach is exponentially hard with respect to the size of $\bar{\a}$ and number of unobserved actions. We thus design an approximate approach in the Expectation-Maximization framework to estimate an unknown plan $\uplan$ that best explains the observation $\obs$. 

To do this, we introduce new parameters to capture ``weights'' of values for each unobserved action. Specifically speaking, assuming there are $X$ unobserved actions in $\obs$, i.e., the number of $\O$s in $\obs$ is $X$, we denote these unobserved actions by $\bar{a}_1,...,\bar{a}_x,...,\bar{a}_X$, where the indices indicate the order they appear in $\obs$. Note that each $\bar{a}_x$ can be any action in $\bar{\a}$. We associate each possible value of $\bar{a}_x$ with a weight, denoted by $\bar{\weight}_{\bar{a}_x,x}$. $\bar{\weight}$ is a $|\bar{\a}|\times X$ matrix, satisfying \[\sum_{o\in\bar{\a}}\bar{\weight}_{o,x}=1,\] where $\bar{\weight}_{o,x}\geq 0$ for each $x$. For the ease of specification, we extend $\bar{\weight}$ to a bigger matrix with a size of $|\bar{\a}|\times M$, denoted by $\weight$, such that $\weight_{o,y} = \bar{\weight}_{o,x}$ if $y$ is the index of the $x$th unobserved action in $\obs$, for all $o\in\bar{\a}$; otherwise, $\weight_{o,y} = 1$ and $\weight_{o',y}=0$ for all $o'\in\bar{\a}\wedge o'\neq o$. Our intuition is to estimate the unknown plan $\uplan$ by selecting actions with the highest weights. We thus introduce the weights to Equation (\ref{prediction}), as shown below,
\begin{eqnarray}\label{withpara}
p(w_{k+j}|w_k)=\prod_{i=1}^{L(w_{k+j})-1}\Big\{\sigma(\one(n(w_{k+j},i+1)= \notag\\ child(n(w_{k+j},i)))\cdot a v_{n(w_{k+j},i)}\cdot b v_{w_k})\Big\},
\end{eqnarray}
where $a=\weight_{w_{k+j},k+j}$ and $b=\weight_{w_k,k}$. We can see that the impact of $w_{k+j}$ and $w_k$ is penalized by weights $a$ and $b$ if they are unobserved actions, and stays unchanged, otherwise (since both $a$ and $b$ equal to 1 if they are observed actions). 

We assume $X\sim Multinomial(\weight_{\cdot,x})$, i.e., $p(X=o) =\weight_{o,x}$, where $\weight_{o,x}\geq 0$ and $\sum_{a\in\bar{\a}}\eta^a = 1$. $P(\uplan|\weight)=\prod_x\weight_{o,x}$

We redefine the objective function as shown below,
\begin{equation}\label{new-way}
\mathcal{F}(\uplan,\weight)=\sum_{k=1}^M\sum_{-c\leq j\leq c,j\neq0} \log p(w_{k+j}|w_k),
\end{equation}
where $p(w_{k+j}|w_k)$ is defined by Equation (\ref{withpara}). The gradient of Equation \ref{new-way} is shown below,
\begin{equation}
\frac{\partial\mathcal{F}}{\partial\weight_{o,x}} = \frac{4c(L(o)-1)}{\weight_{o,x}}.
\end{equation}
The only parameters needed to be updated are $\weight$, which can be easily done by gradient descent, as shown below,
\begin{equation}\label{update}
\weight_{o,x} = \weight_{o,x} + \delta\frac{\partial\mathcal{F}}{\partial\weight_{o,x}},
\end{equation}
if $x$ is the index of unobserved action in $\obs$; otherwise, $\weight_{o,x}$ stays unchanged, i.e., $\weight_{o,x}=1$. Note that $\delta$ is a learning constant. 

With Equation (\ref{update}), we can design an EM algorithm by repeatedly sampling an unknown plan according to $\weight$ and updating $\weight$ based on Equation (\ref{update}) until reaching convergence (e.g., a constant number of repetitions is reached).

\subsection{Overview of our {\dup} approach}
An overview of our {\dup} algorithm is shown in Algorithm \ref{dup}. In Step 2 of Algorithm \ref{dup}, we initialize $\weight_{o,k}=1/M$ for all $o\in\bar{\a}$, if $k$ is an index of unobserved actions in $\obs$; and otherwise, $\weight_{o,k}=1$ and $\weight_{o',k}=0$ for all $o'\in\bar{\a}\wedge o'\neq o$. In Step 4, we view $\weight_{\cdot,k}$ as a probability distribution, and sample an action from $\bar{\a}$ based on $\weight_{\cdot,k}$ if $k$ is an unobserved action index in $\obs$. In Step 5, we only update $\weight_{\cdot,k}$ where $k$ is an unobserved action index. In Step 6, we linearly project all elements of the updated $\weight$ to between 0 and 1, such that we can do sampling directly based on $\weight$ in Step 4. In Step 8, we simply select $\bar{a}_x$ based on \[\bar{a}_x=\arg\max_{o\in\bar{\a}}\weight_{o,x},\] for all unobserved action index $x$.
\begin{algorithm}[!ht]
\caption{Framework of our {\dup} algorithm}\label{dup}
\textbf{Input:} plan library $\lib$, observed actions $\obs$ \\
\textbf{Output:} plan $\uplan$
\begin{algorithmic}[1]
\STATE learn vector representation of actions
\STATE initialize $\weight_{o,k}$ with $1/M$ for all $o\in\bar{\a}$, when $k$ is an unobserved action index
\WHILE{the maximal number of repetitions is not reached} 
	\STATE sample unobserved actions in $\obs$ based on $\weight$
	\STATE update $\weight$ based on Equation (\ref{update})
	\STATE project $\weight$ to [0,1]
\ENDWHILE
\STATE select actions for unobserved actions with the largest weights in $\weight$
\RETURN $\uplan$
\end{algorithmic}
\end{algorithm}

Our {\dup} algorithm framework belongs to a family of policy gradient
algorithms, which have been successfully applied to complex problems,
e.g., robot control \cite{cof/nips/ng03}, natural language processing
\cite{cof/acl/Branavan12}. Our formulation is unique in how it
recognizes plans, in comparison to the existing methods in the planning
community.

Note that our current study shows that even direct application of word
vector learning methods provide competitive performance for plan
completion tasks. We believe we can further improve the performance by
using the planning specific structural information in the EM phase. In
other words, if we are  provided with additional planning structural
information as input, we can exploit the structural information to
filter candidate plans to be recognized in the EM procedure.

\section{Our {\rnnplanner} approach}
Instead of using the EM-style framework, in this section we present another approach which is based on Recurrent Neural Networks (RNNs), specifically Long Short-term Memory networks (LSTMs), with the distributed representations of actions introduced in Section \ref{actionvector}. LSTM is a specific kind of RNN that works by leveraging long-short term contexts. As we model our plan recognition problem as an action-sequence generation problem, our aim of exploring RNN-LSTM architecture is to leverage longer-horizon of action contexts to help improve the accuracy of generating new actions based on previously observed or generated actions. We will first introduce the RNN-LSTM architecture, and then introduce our {\rnnplanner} model. 



\subsection{The RNN Model}
Specifically, the RNN architecture can be defined in the following way. Given an input action $x_t$ at the step $t$, RNN accepts it with weighted connections to $N$ hidden layers that are stacked together. And from the hidden layer stack, there is a connection to the output layer $y_t$, as well as a cyclic weighted connection going into the hidden layer stack. And if we unroll this RNN cell along $T$ steps, it could accept an action input sequence $\mathbf{x} = (x_1...,x_T)$, and compute a sequence of hidden states $\mathbf{h}=(h_1^N,h_2^N,...,h_T^N)$. For each of these hidden states $h_t$ ($1 \leq t \leq T$), it contributes to predicting the next step output $y_{t+1}$, and thus RNN computes an output vector sequence $\mathbf{y}=(y_1,...,y_T)$, by concatenating outputs from all steps together.

\begin{figure}[!ht]
\centerline{\includegraphics[width=0.99\textwidth]{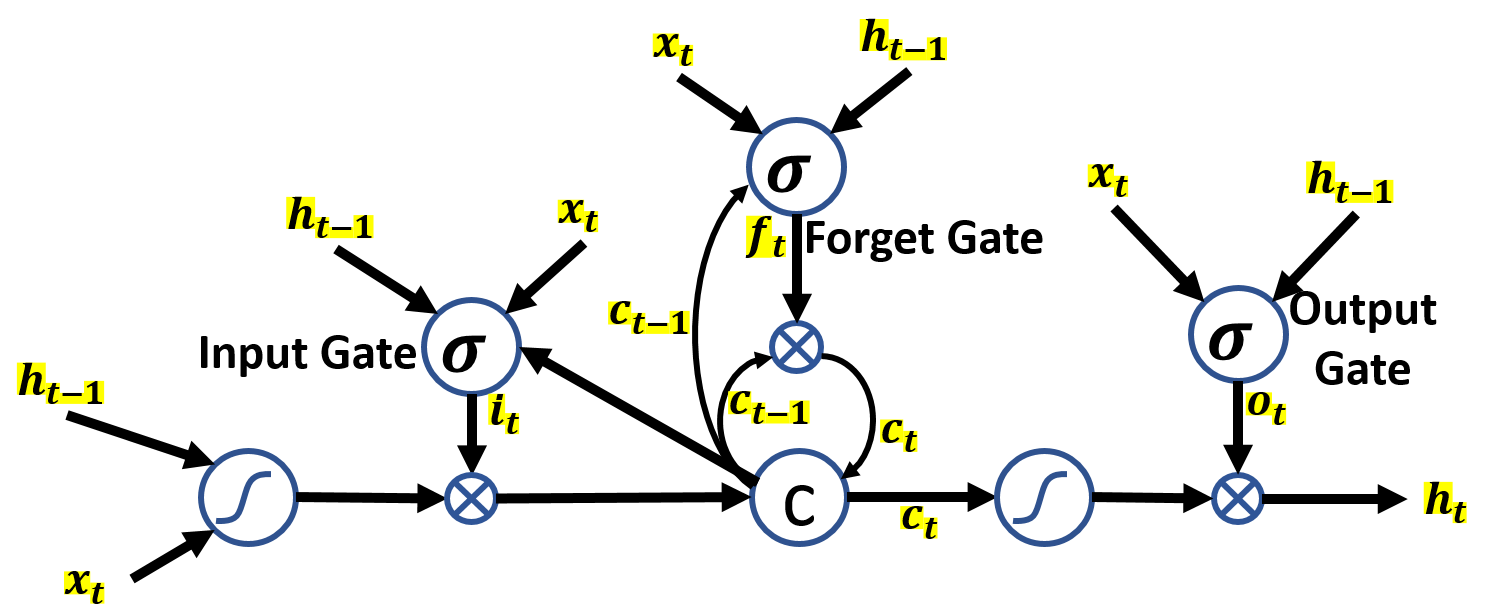}}
\caption{Long Short-term Memory (LSTM) cell}
\label{lstm_cell}
\end{figure}

Given an input sequence $\mathbf{x}$, an RNN model could predict an output sequence $\mathbf{y}$, in which output $y_t$ at each step depends on the input $x_t$ at that step, and the hidden state at the previous step. The RNN could also be utilized to directly generate, in principle, infinitely long future outputs (actions), given a single input $x_t$. The sequence of future actions could be generated by directly feeding the output $y_t$ at a step $t$, to the input $x_{t+1}$ at the next step $t+1$. This way, RNN ``assumes'' what it predicts that would happen at next step is reliable ($y_t = x_{t+1}$). As for training the RNN as a sequence generation model, we could utilize $y_t$ to parameterize a predictive distribution $P(x_{t+1}|y_t)$ over all of the possible next inputs $x_{t+1}$, and thus we could minimize the loss:

\begin{equation}\label{probability-of-sequence}
\mathcal{L}(\mathbf{x})=-\sum_{t=1}^T log P(x_{t+1}|y_t).
\end{equation} where $T$ is the number of steps of an observed plan trace, $x_{t+1}$ is the observed action at step $t+1$, and $y_t$ is the output at step $t$ as well as the prediction of what would happen at step $t+1$. To estimate $y_t$ based on $x_1,\ldots,x_t$, we exploit the Long Short-term Memory (LSTM) model, which has been demonstrated effective on generating sequences \cite{DBLP:journals/corr/Graves13,DBLP:conf/nips/ShiCWYWW15}, to leverage long term information prior to $x_t$ and predict $y_t$ based on current input $x_t$. We can thus rewrite Equation (\ref{probability-of-sequence}) as:

\begin{equation}
\mathcal{L}(\mathbf{x})=-\sum_{t=1}^T log P(x_{t+1}|y_t)\mathrm{LSTM}(y_t|x_{1:t}),
\label{loss}
\end{equation}
where $\mathrm{LSTM}(y_t|x_{1:t})$ indicates the LSTM model estimates $y_t$ based on current input $x_t$ and memories of previous input prior to $x_t$. The framework of LSTM \cite{DBLP:journals/corr/Graves13,DBLP:conf/nips/ShiCWYWW15} is shown in Figure \ref{lstm_cell}, where $x_t$ is the $t$th input, $h_t$ is the $t$th hidden state. $i_t$, $f_t$, $o_t$ and $c_t$ are the $t$th \emph{input gate}, \emph{forget gate}, \emph{output gate}, \emph{cell} and \emph{cell input} activation vectors, respectively, whose dimensions are the same as the hidden vector $h_t$.

LSTM is implemented by the following functions:
\begin{eqnarray}
i_t &=& \sigma(W_{xi}x_t+W_{hi}h_{t-1}+W_{ci} c_{t-1}+b_i) \\
f_t &=& \sigma(W_{xf}x_t+W_{hf}h_{t-1}+W_{cf} c_{t-1}+b_f) \\
c_t &=& f_t \circ c_{t-1} + i_t \circ \tanh(W_{xc}x_t+W_{hc}h_{t-1}+b_c) \\
o_t &=& \sigma(W_{xo}x_t+W_{ho}h_{t-1}+W_{co} c_t+b_o) \\
h_t &=& o_t \circ \tanh(c_t)
\label{lstm_cell_formula}
\end{eqnarray}
where $\circ$ indicates the Hadamard product, $\sigma$ is the logistic sigmoid function, $W_{xi}$ is an input-input gate matrix,  $W_{hi}$ is a hidden-input gate matrix, $W_{ci}$ is a cell-input gate matrix, $W_{xf}$ is an input-forget gate matrix, $W_{hf}$ is a hidden-forget gate matrix, $W_{cf}$ is a cell-forget gate matrix, $W_{xc}$ is an input-cell gate matrix, $W_{hc}$ is a hidden-cell gate matrix,  $W_{xo}$ is an input-output gate matrix, $W_{ho}$ is a hidden-output gate matrix, $W_{co}$ is a cell-output gate matrix. $b_i$, $b_f$, $b_c$, and $b_o$ are bias terms. Note that the matrices from cell to gate vectors (i.e., $W_{ci}$, $W_{cf}$ and $W_{co}$) are diagonal, such that each element $e$ in each gate vector only receives input of element $e$ of the cell vector. The major innovation of LSTM is its memory cell $c_t$ which essentially acts as an accumulator of the state information. $c_t$ is accessed, written and cleared by self-parameterized controlling gates, i.e., \emph{input, forget, output} gates. Each time a new input $x_t$ comes, its information is accumulated to the memory cell if the input gate $i_t$ is activated. The past cell status $c_{t-1}$ could be forgotten in this process if the forget gate $f_t$ is activated. Whether the latest cell output $c_t$ is propagated to the final state $h_t$ is further controlled by the output gate $o_t$. The benefit of using the memory cell and gates to control information flow is the gradient is trapped in the cell and prevented from vanishing too quickly.

\subsection{Discovering Underlying Plans with the RNN Model}
\begin{figure}[!ht]
\centerline{\includegraphics[width=0.99\textwidth]{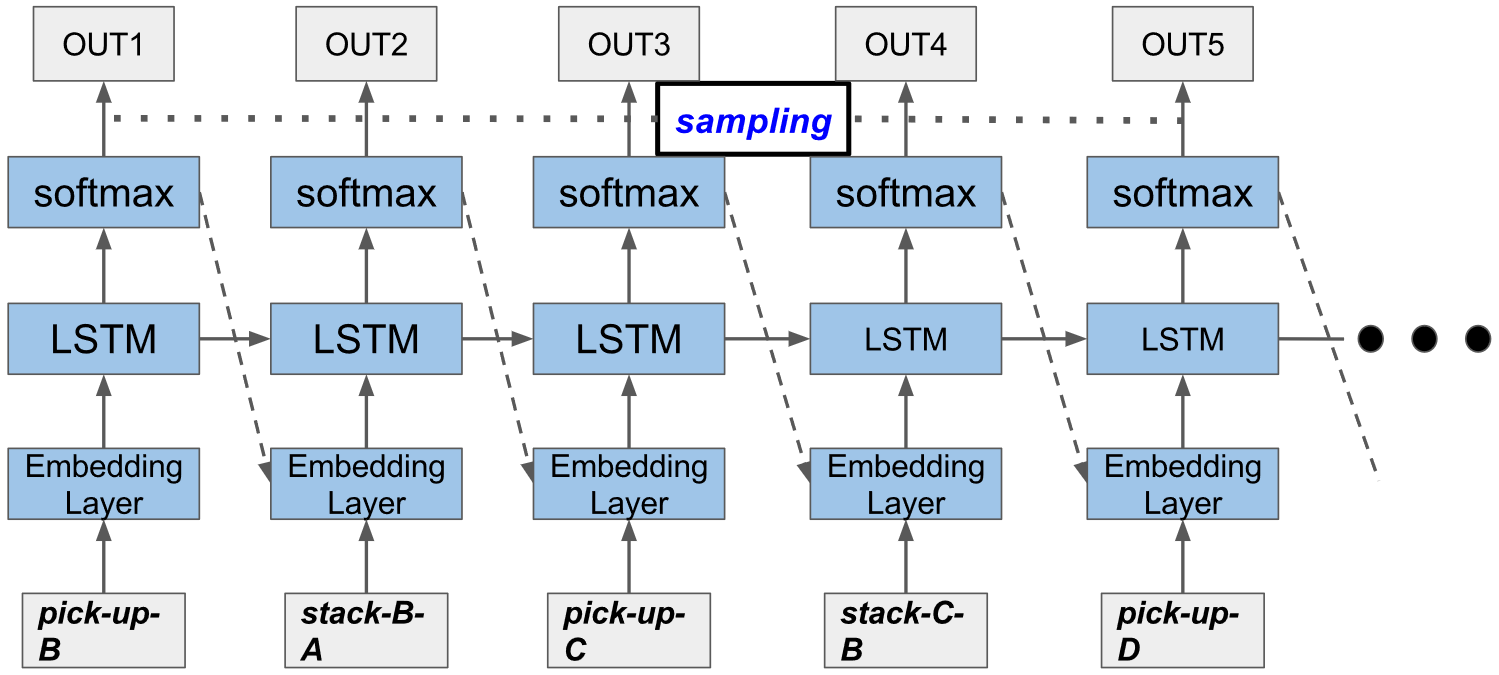}}
\caption{The framework of our {\rnnplanner} approach (with one hidden layer)}
\label{arch_rnnplanner}
\end{figure}

With the distributed representations of actions addressed in Section 3, we can view each plan in the plan library as a sequence of actions, and the plan library as a set of action sequences which can be utilized to train the RNN model. The framework of RNN with sequences of actions can be seen from Figure \ref{arch_rnnplanner}. The bottom row in Figure \ref{arch_rnnplanner} is an example action sequence ``pick-up-B, stack-B-A, pick-up-C, stack-C-B, pick-up-D, ...'', which corresponds to an input sequence $\mathbf{x}$. Once an action among the bottom row is fed into the RNN, that action is assigned with an index, and an embedding layer is trained to find a vector representation based on that index. The action vector from the embedding layer is the feature that can be used by the LSTM cell. How the LSTM cell works has been explained in Equation \ref{lstm_cell_formula}. Similar to a classic RNN cell, the LSTM cell feeds its output to both itself as a hidden state, and the softmax layer to obtain a probability distribution over all actions in the action vocabulary $\a$. From the perspective of the LSTM cell at the next step, it receives a hidden state from the previous step $h_{t-1}$, an action vector at the current step $x_t$. To obtain the index of most possible action, our model samples over the action distribution output from softmax layer. That retrieved index could be mapped to an action in the vocabulary $\bar{\a}$. 

The top row in Figure \ref{arch_rnnplanner} is the output sequence, which is denoted by ``OUT1, OUT2, OUT3, OUT4, OUT5, ...'', which corresponds to the estimated sequence ``$y_1$, $y_2$, $y_3$, $y_4$, $y_5$, $\ldots$'' in Equation (\ref{loss}).

Note that we exploit the dotted arrow to indicate two folds of meanings in Figure \ref{arch_rnnplanner}. When training the RNN model, the one pointed by the head of the dotted row (the embedding of input) is used to compute the cross entropy error with the output at tail (output of LSTM cell at the previous step), and next-step observation as the input at the head, to train the model. When using the trained RNN model to discover unknown actions, the model ``imagines'' what it predicts is the real next input, and takes it to continue its prediction. Thus the one pointed by the head is copied and identical to the one denoted by the tail. For example, the embedding of ``stack-B-A'' is copied from the prediction vector of ``OUT1'' if the input ``stack-B-A'' was unknown. In addition, the arrows between each of two LSTM cells shows the unrolling of a LSTM cell. The horizontal dashed line suggests that we obtain the action output at each step, by sampling from probability distribution, provided by the softmax layer.

With the trained RNN model, we can discover underlying actions by simply exploiting the RNN model to generate unknown actions based on observed or already discovered actions. For example, given the observation:
\begin{center}
\emph{pick-up-B  $\O$ unstack-D-C put-down-D  $\O$ stack-C-B $\O$ $\O$}
\end{center}
we can generate the first $\O$ based on \emph{pick-up-B}, the second $\O$ based on actions from \emph{pick-up-B} to \emph{put-down-D}, the third $\O$ based on actions from \emph{pick-up-B} to \emph{stack-C-B}, and the last $\O$ based on all previously actions (including generated actions at where there was a $\O$).

\section{Experiments}
In this section, we evaluate our {\dup} and {\rnnplanner} algorithms in three planning domains from International Planning Competition, i.e., blocks$^1$, depots\footnote{http://www.cs.cmu.edu/afs/cs/project/jair/pub/volume20/long03a-html/JAIRIPC.html}, and driverlog$^2$. To generate training and testing data, we randomly created 5000 planning problems for each domain, and solved these planning problems with a planning solver, such as FF\footnote{https://fai.cs.uni-saarland.de/hoffmann/ff.html}, to produce 5000 plans. 

We define the accuracy of our {\dup} and {\rnnplanner} algorithms as follows. For each unobserved action $\bar{a}_x$, {\dup} and {\rnnplanner} suggest a set of possible actions $S_x$ which have the highest value of $\weight_{\bar{a}_x,x}$ for all $\bar{a}_x\in\bar{\a}$. If $S_x$ covers the \emph{truth} action $a_{truth}$, i.e., $a_{truth}\in S_x$, we increase the number of correct suggestions by 1. We thus define the accuracy $acc$ as shown below:
\[acc = \frac{1}{T}\sum_{i=1}^T\frac{\#\langle correct\textrm{-}suggestions\rangle_i}{K_i},\]
where $T$ is the size of testing set, $\#\langle correct\textrm{-}suggestions\rangle_i$ is the number of correct suggestions for the $i$th testing plan, $K_i$ is the number of unobserved actions in the $i$th testing plan. We can see that the accuracy $acc$ may be influenced by $S_x$. We will test different size of $S_x$ in the experiment.

\begin{table}[!th]
\centering
\caption{Features of datasets}\label{dataset}
\begin{tabular}{|l|l|l|l|}
\hline
domain   & \#plan & \#word & \#vocabulary \\\hline
blocks     & 5000 & 292250 & 1250\\\hline
depots     & 5000 & 209711 & 2273\\\hline
driverlog     & 5000 & 179621 & 1441\\\hline
\end{tabular}
\end{table}

State-of-the-art plan recognition approaches with plan libraries as input aim at finding a plan from plan libraries to best explain the observed actions \cite{DBLP:conf/ijcai/GeibS07}, which we denote by {\tt MatchPlan}. We develop a {\tt MatchPlan} system based on the idea of \cite{DBLP:conf/ijcai/GeibS07} and compare our {\dup} algorithm to {\tt MatchPlan} with respect to different percentages of unobserved actions $\xi$ and different sizes of suggestion or recommendation set $S_x$. Another baseline is action-models based plan recognition approach \cite{cof/ijcai/Ramirez09} (denoted by {\tt PRP}, short for Plan Recognition as Planning). Since we do not have action models as input in our {\dup} algorithm, we exploited the action model learning system {\tt ARMS} \cite{journal/aij/Yang07} to learn action models from the plan library and feed the action models to the {\tt PRP} approach. We call this hybrid plan recognition approach {\tt ARMS+PRP}. To learn action models, {\tt ARMS} requires state information of plans as input. We thus added extra information, i.e., initial state and goal of each plan in the plan library, to {\tt ARMS+PRP}. In addition, {\tt PRP} requires as input a set of candidate goals $\mathcal{G}$ for each plan to be recognized in the testing set, which was also generated and fed to {\tt PRP} when testing. In summary, the hybrid plan recognition approach {\tt ARMS+PRP} has more input information, i.e., initial states and goals in plan library and candidate goals $\mathcal{G}$ for each testing example, than our {\dup} approach.

To evaluate DUP, we compared it with several baselines that we elaborated above, i.e., {\tt MatchPlan}, and {\tt ARMS+PRP}. We randomly divided the plans into ten folds, with 500 plans in each fold. We ran our {\dup} algorithm ten times to calculate an average of accuracies, each time with one fold for testing and the rest for training. In the testing data, we randomly removed actions from each testing plan (i.e., $\obs$) with a specific percentage $\xi$ of the plan length. Features of datasets are shown in Table \ref{dataset}, where the second column is the number of plans generated, the third column is the total number of words (or actions) of all plans, and the last column is the size of vocabulary used in all plans. 
To evaluate our {\rnnplanner} algorithm, we directly compared {\rnnplanner} to {\dup}. 


\begin{figure}[!ht]
\centerline{\includegraphics[width=0.99\textwidth]{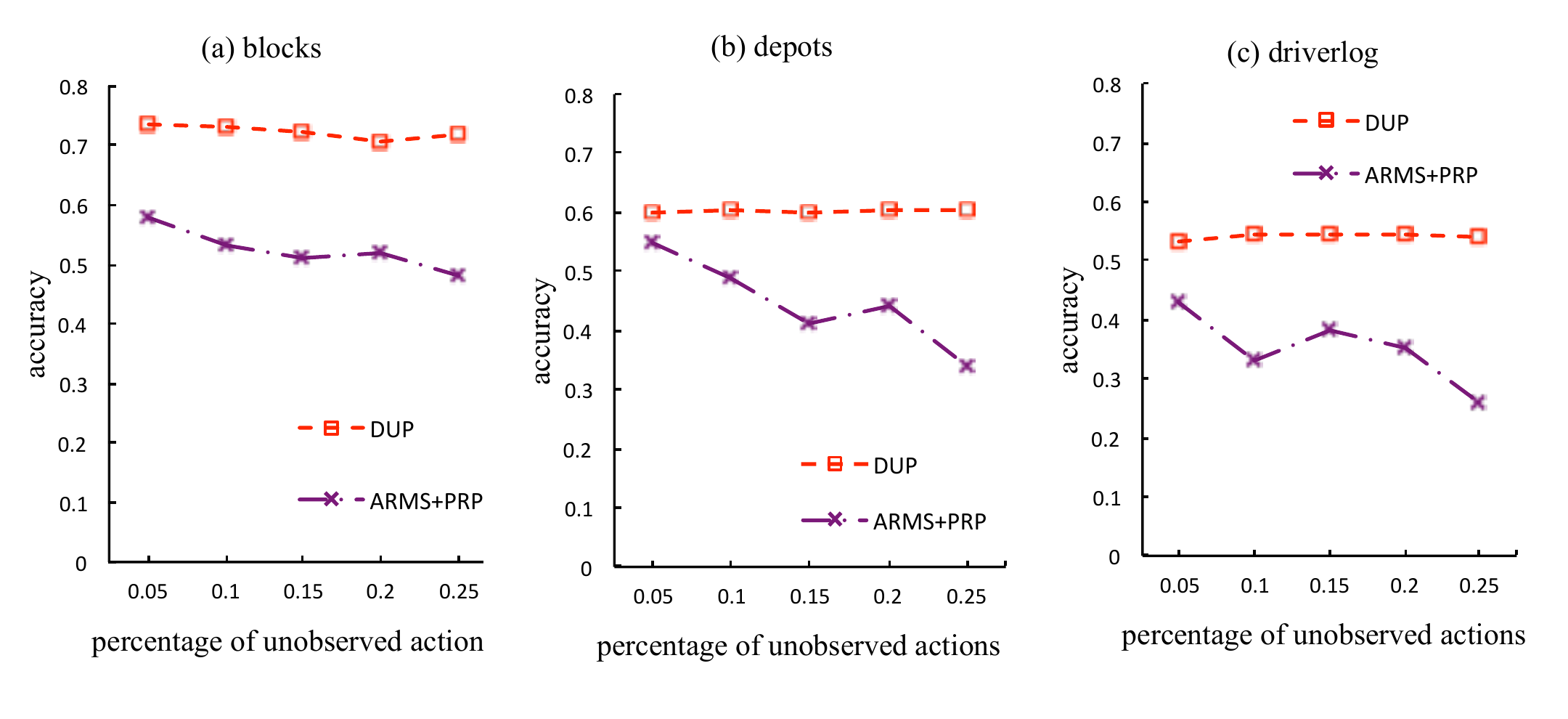}}
\caption{Accuracies of {\dup} and {\tt ARMS+PRP} with respect to different percentage of unobserved actions}
\label{prp-percentage}
\end{figure}
\begin{figure}[!ht]
\centerline{\includegraphics[width=0.99\textwidth]{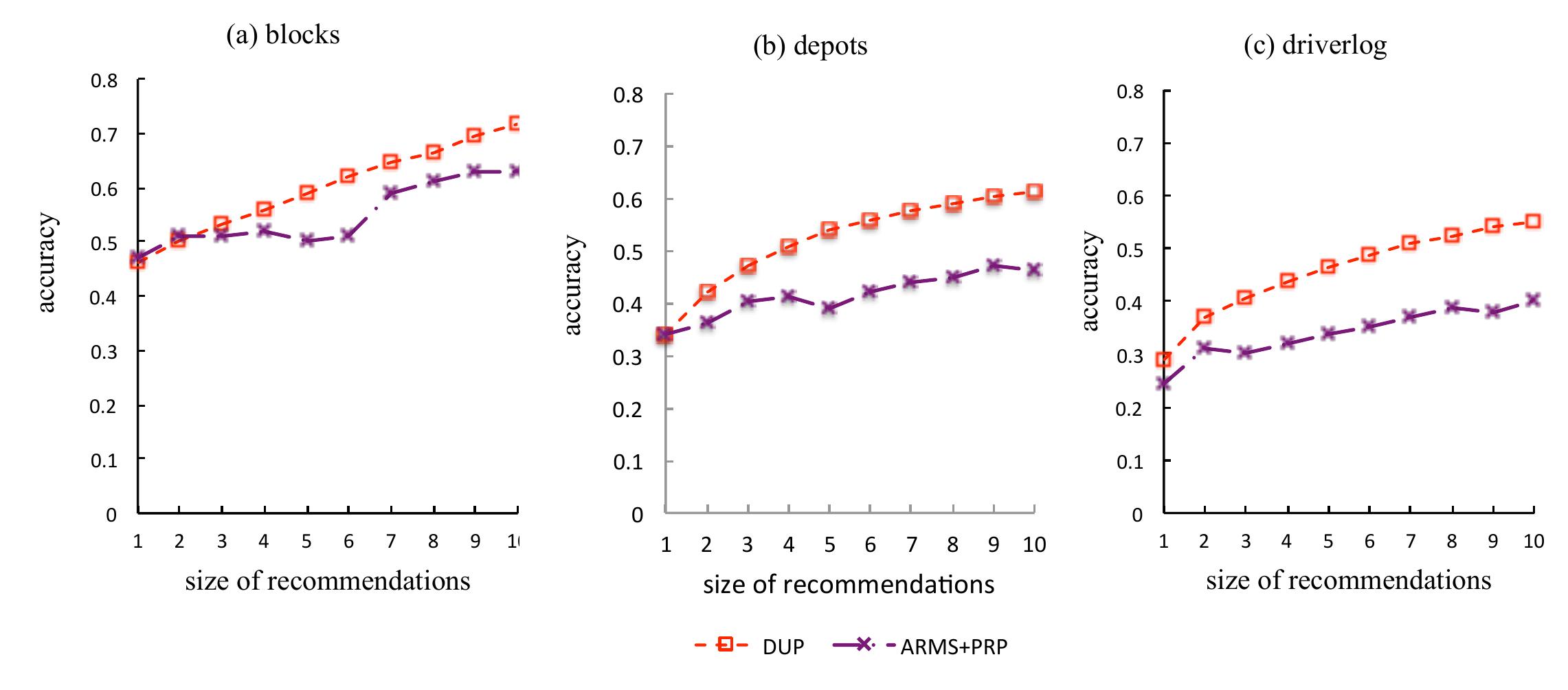}}
\caption{Accuracies of {\dup} and {\tt ARMS+PRP} with respect to different size of recommendations}
\label{prp}
\end{figure}

\subsection{Comparison between {\dup} and {\tt ARMS+PRP}}
We first compare our {\dup} algorithm to {\tt ARMS+PRP} to see the advantage of {\dup}. We varied the percentage of unobserved actions and the size of recommended actions to see the change of accuracies, respectively. The results are shown below. 

\subsubsection{Varying Percentage of Unobserved Actions}
In this experiment we would like to see the change of accuracies of both our {\dup} algorithm and {\tt ARMS+PRP} with respect to $\xi$ in $\obs$. We set the window of training context $c$ in Equation (\ref{skip-gram}) to be three, the number of iterations in Algorithm \ref{dup} to be 1500, the size of recommendations to be ten, and the learning constant $\delta$ in Equation (\ref{update}) to be 0.1.  For {\tt ARMS+PRP}, we generated 20 candidate goals for each testing example including the ground-truth goal which corresponds to the ground-truth plan to be recognized. The results are shown in Figure \ref{prp-percentage}. 

From Figure \ref{prp-percentage}, we can see that in all three domains, the accuracy of our {\dup} algorithm is generally higher {\tt ARMS+PRP}, which verifies that our {\dup} algorithm can indeed capture relations among actions better than the model-based approach {\tt ARMS+PRP}. The rationale is that we explore global plan information from the plan library to learn a ``shallow'' model (distributed representations of actions) and use this model with global information to best explain the observed actions. While {\tt ARMS+PRP} tries to leverage global plan information from the plan library to learn action models and uses the models to recognize observed actions, it enforces itself to extract ``exact'' models represented by planning models which are often with noise. When feeding those noisy models to {\tt PRP}, since {\tt PRP} that uses planning techniques to recognize plans is very sensitive to noise of planning models, the recognition accuracy is lower than {\dup}, even though {\tt ARMS+PRP} has more input information (i.e., initial states and candidate goals) than our {\dup} algorithm.

\begin{figure*}[!ht]
\centerline{\includegraphics[width=0.99\textwidth]{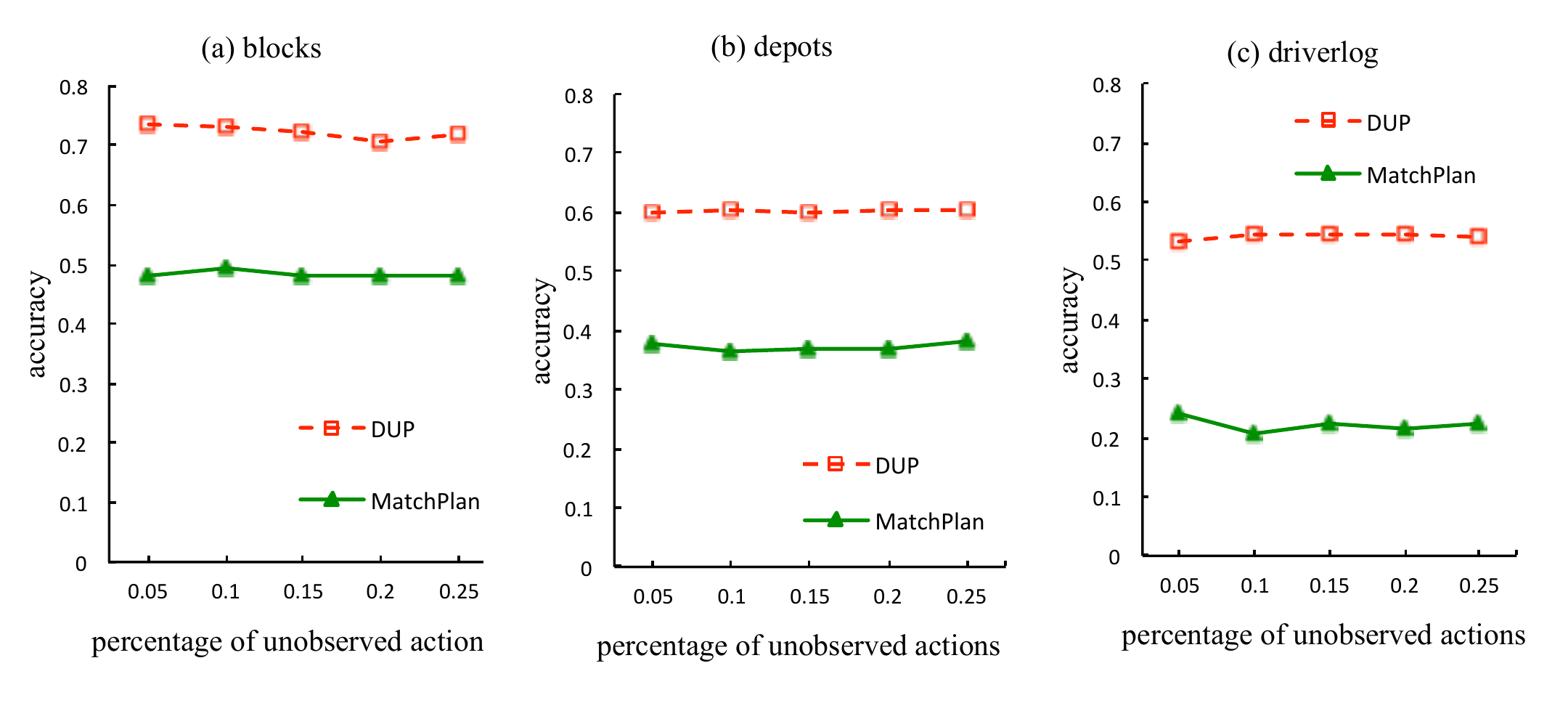}}
\caption{Accuracies of {\dup} and {\tt MatchPlan} with respect to different percentage of unobserved actions}
\label{DUP2MatchPlan-percentage}
\end{figure*}

Looking at the changes of accuracies with respect to the percentage of
unobserved actions, we can see that our {\dup} algorithm performs
fairly well even when the percentage of unobserved action reaches
25\%. In contrast,  {\tt ARMS+PRP} is sensitive to the percentage of unobserved
actions, i.e., the accuracy goes down when more actions are
unobserved. This is because the noise of planning models induces more
uncertain information, which harms the recognition accuracy, when the
percentage of unobserved actions becomes larger.  Comparing accuracies
of different domains, we can see that our {\dup} algorithm functions
better in the \emph{blocks} domain than the other two domains. This is
because the ratio of \#word over \#vocabulary in the \emph{blocks}
domain is much larger than the other two domains, as shown in Table
\ref{dataset}. We would conjecture that increasing the ratio could
improve the accuracy of {\dup}. From Figure \ref{prp-percentage} (as well as Figure \ref{DUP2MatchPlan-percentage}), we can see that it appears that the accuracy of DUP is not 
affected by increasing percentages of unobserved actions. The rationale is (1) the percentage of unobserved actions is low, 
less than 25\%, i.e., 
there is at most one unobserved action over four continuous actions; (2) the window size of 
context in DUP is set to be 3, which ensures that DUP generally has ''stable'' context information to estimate 
the unobserved action when the percentage of unobserved actions is less than 25\%, resulting in the stable accuracy
in Figure \ref{prp-percentage} (likewise for Figure \ref{DUP2MatchPlan-percentage}).

\subsubsection{Varying Size of Recommendation Set}
We next evaluate the performance of our {\dup} algorithm with respect to the size of recommendation set $S_x$. We evaluate the influence of the recommendation set by varying the size from 1 to 10. The size of recommendation set is much smaller than the complete set. For example, the size of complete set in the blocks domain is 1250 (shown in Table 1). It is less than 1\% even though we recommend 10 actions for each unobserved action. We set the context window $c$ used in Equation (\ref{skip-gram}) to be three, the percentage of unobserved actions to be 0.25, and the learning constant $\delta$ in Equation (\ref{update}) to be 0.1. For {\tt ARMS+PRP}, the number of candidate goals for each testing example is set to 20. {\tt ARMS+PRP} aims to recognize plans that are optimal with respect to the cost of actions. We relax {\tt ARMS+PRP} to output $|S_x|$ optimal plans, some of which might be suboptimal. The results are shown in Figure \ref{prp}.

From Figure \ref{prp}, we find that accuracies of the three approaches generally become larger when the size of the recommended set increases in all three domains. This is consistent with our intuition, since the larger the recommended set is, the higher the possibility for the \emph{truth} action to be in the recommended set. We can also see that the accuracy of our {\dup} algorithm are generally larger than {\tt ARMS+PRP} in all three domains, which verifies that our {\dup} algorithm can indeed better capture relations among actions and thus recognize unobserved actions better than the model-learning based approach {\tt ARMS+PRP}. The reason is similar to the one given for Figure \ref{prp-percentage} in the previous section. That is, the ``shallow'' model learnt by our {\dup} algorithm is better for recognizing plans than both the ``exact'' planning model learnt by {\tt ARMS} for recognizing plans with planning techniques. Furthermore, the advantage of {\dup} becomes even larger when the size of recommended action set increases, which suggests our vector representation based learning approach can better capture action relations when the size of recommended action set is larger. The possibility of actions correctly recognized by {\dup} becomes much larger than {\tt ARMS+PRP}  when the size of recommendations increases.

\begin{figure*}[!ht]
\centerline{\includegraphics[width=0.99\textwidth]{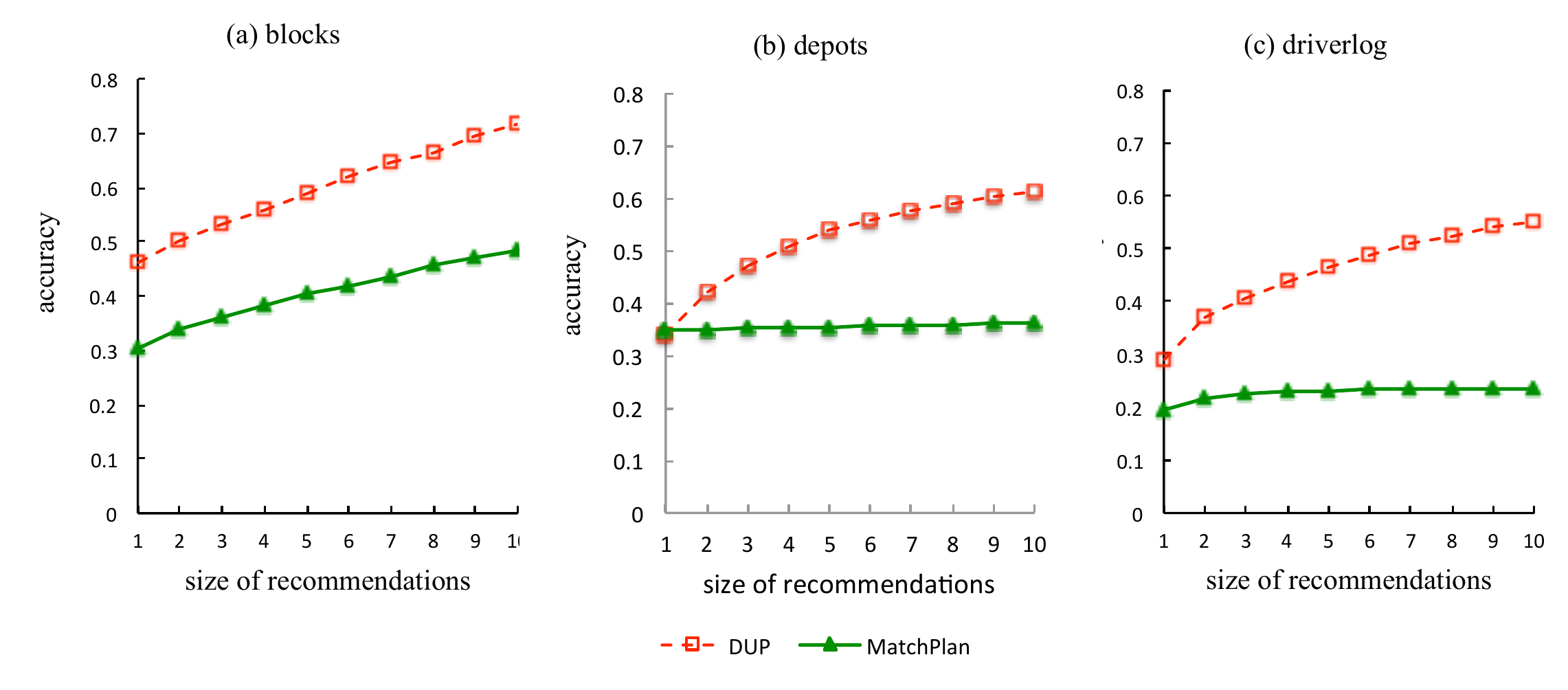}}
\caption{Accuracies of {\dup} and {\tt MatchPlan} with respect to different size of recommendations}
\label{DUP2MatchPlan}
\end{figure*}
\begin{figure*}[!ht]
\centerline{\includegraphics[width=0.99\textwidth, height=6.8cm]{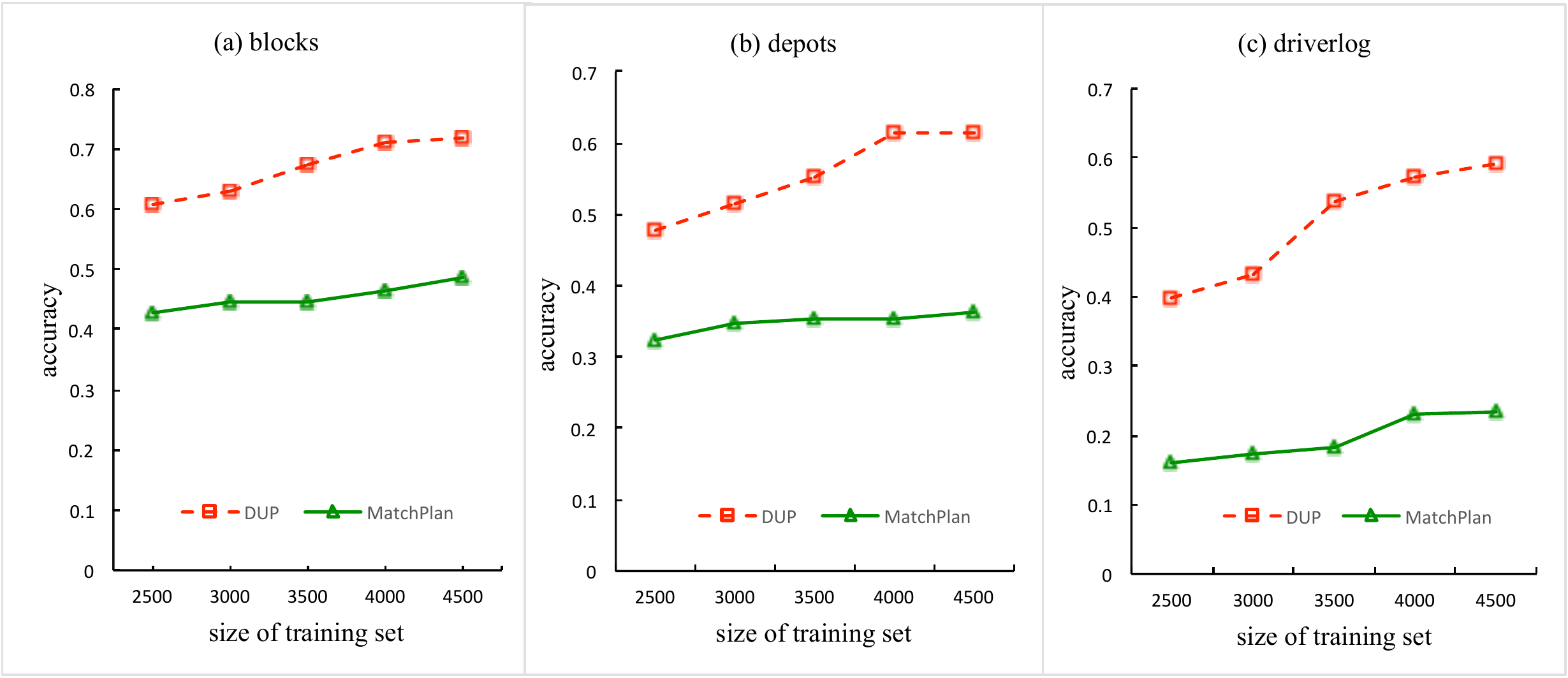}}
\caption{Accuracies of {\dup} and {\tt MatchPlan} with respect to different size of training set}
\label{DUP2MatchPlan-size}
\end{figure*}


\subsection{Comparison between {\dup} and {\tt MatchPlan}}
In this experiment we compare {\dup} to {\tt MatchPlan} which is built based on the idea of \cite{DBLP:conf/ijcai/GeibS07}. Likewise we varied the percentage of unobserved actions and the size of recommended actions to see the change of accuracies of both algorithms.  The results are exhibited below.
\subsubsection{Varying Percentage of Unobserved Actions}
To compare our {\dup} algorithm with {\tt MatchPlan} with respect to different percentage of unobserved actions, we set the window of training context $c$ in Equation (\ref{skip-gram}) of {\dup} to be three, the number of iterations in Algorithm \ref{dup} to be 1500, the size of recommendations to be ten, and the learning constant $\delta$ in Equation (\ref{update}) to be 0.1.
To make fair the comparison (with {\tt MatchPlan}), we set the matching window of {\tt MatchPlan} to be three, the same as the training context $c$ of {\dup}, when searching plans from plan libraries $\lib$. In other words, to estimate an unobserved action $\bar{a}_x$ in $\obs$, {\tt MatchPlan} matches previous three actions and subsequent three actions of $\bar{a}_x$ to plans in $\lib$, and recommends ten actions with the maximal number of matched actions, considering observed actions in the context of $\bar{a}_x$ and actions in $\lib$ as a successful matching. The results are shown in Figure \ref{DUP2MatchPlan-percentage}.

From Figure \ref{DUP2MatchPlan-percentage}, we find that the accuracy of {\dup} is much better than {\tt MatchPlan}, which indicates that our {\dup} algorithm can better learn knowledge from plan libraries than the local matching approach {\tt MatchPlan}. This is because we take advantage of global plan information of the plan library when learning the ``shallow'' model, i.e., distributed representations of actions, and the model with global information can best explain the observed actions.
In contrast, {\tt MatchPlan} just utilizes local plan information when matching the observed actions to the plan library, which results in lower accuracies. Looking at all three different domains, we can see that both algorithms perform the best in the \emph{blocks} domain. The reason is similar to the one provided in the last subsection (for Figure \ref{prp-percentage}), i.e., the number of words over the number of vocabulary in the \emph{blocks} domain is relatively larger than the other two domains, which gives us the hint that it is possible to improve accuracies by increasing the ratio of the number of words over the number of vocabularies.

\subsubsection{Varying Size of Recommendation Set}
Likewise, we also would like to evaluate the change of accuracies when increasing the size of recommended actions. We used the same experimental setting as done by previous subsection. That is, we set the window of training context $c$ of {\dup} to be three, the learning constant $\delta$ to be 0.1, the number of iterations in Algorithm \ref{dup} to be 1500, the matching window of {\tt MatchPlan} to be three. In addition, we fix the percentage of unobserved actions to be 0.25. The results are shown in Figure \ref{DUP2MatchPlan}.
 
We can observe that the accuracy of our {\dup} algorithm are generally larger than {\tt MatchPlan} in all three domains in Figure \ref{DUP2MatchPlan}, which suggests that our {\dup} algorithm can indeed better capture relations among actions and thus recognize unobserved actions better than the matching approach {\tt MatchPlan}. The reason behind this is similar to previous experiments, i.e., the global information  captured from plan libraries by {\dup} can indeed better improve accuracies than local information exploited by {\tt MatchPlan}. In addition, looking at the trends of the curves of both {\dup} and {\tt MatchPlan}, we can see the performance of {\dup} becomes much better than {\tt MatchPlan} when the size of recommendations increases. This indicates the influence of global information becomes much larger when the size of recommendations increasing. In other words, larger size of recommendations provides better chance for ``shallow'' models learnt by {\dup} to perform better. 

\subsubsection{Varying Size of Training Set}
To see the effect of size of training set, we ran both {\dup} and {\tt MatchPlan} with different size of training set. We used the same setting as done by last subsection except fixing the size of recommendations to be 10, when running both algorithms. We varied the size of training set from 2500 to 4500. The results are shown in Figure \ref{DUP2MatchPlan-size}.

We observed that accuracies of both {\dup} and {\tt MatchPlan} generally become higher when the size of training set increases. This is consistent with our intuition, since the larger the size of training set is, the richer the information is available for improving the accuracies. Comparing the curves of {\dup} and {\tt MatchPlan}, we can see that {\dup} performs much better than {\tt MatchPlan}. This further verifies the benefit of exploiting global information of plan libraries when learning the shallow models as done by {\dup}. 

\begin{figure}[!ht]
\centerline{\includegraphics[width=0.65\textwidth]{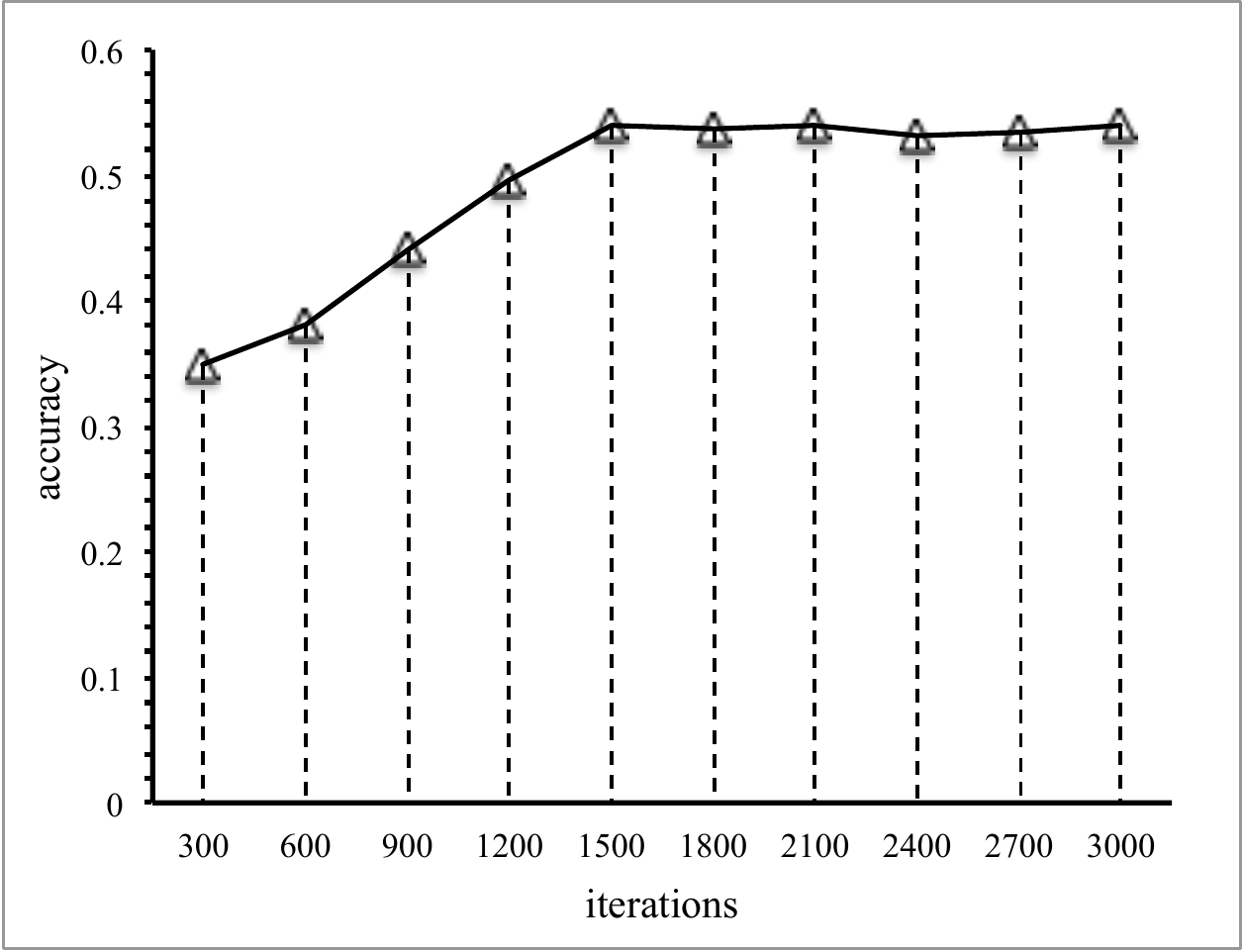}}
\caption{Accuracy with respect to different number of iterations in the blocks domain}
\label{iterations}
\end{figure}

\subsection{Accuracy w.r.t. Iterations}
In the previous experiments, we set the number of iterations in Algorithm \ref{dup} to be 1500. In this experiment, we would like to see the influence of iterations of our {\dup} algorithm when running the EM-style procedure. We changed the number of iterations from 300 to 3000 to see the trend of accuracy. We exhibit the experimental results in the \emph{blocks} domain (the results of the other two domains are similar) in Figure \ref{iterations}.

From Figure \ref{iterations}, we can see the accuracy becomes higher at the beginning and stays flat when reaching the size of 1500. This exhibits that the EM procedure converges and has stable accuracies after the iteration reaches 1500. Similar results can also be found in the other two domains.

\subsection{Comparison between {\rnnplanner} and {\dup}}
In this section we compare {\rnnplanner} with {\dup} to see the change of performance with respect to different distributions of missing actions in the underlying plans to be discovered. In this experiment, we are interested in evaluating the performance on consecutive missing actions in the underlying plans since these scenarios often exist in many applications such as surveillance \cite{surveillanceExample}. We first test the performance of both {\rnnplanner} and {\dup} in discovering underlying plans with only consecutive missing actions in the ``middle'' of the plans, i.e., actions are not missing at the end or in the front, which indicates missing actions can be inferred from both previously and subsequently observed actions. Then we evaluate both {\rnnplanner} and {\dup} in discovering underlying plans with only consecutive missing actions at the end of the plans, which indicates missing actions can only be inferred from previously observed actions. After that, we also evaluate the performance of our {\rnnplanner} and {\dup} approaches with respect to the size of recommendation set. In the following subsections, we present the experimental results regarding those three aspects.

\subsubsection{Performance with missing actions in the middle}
\begin{figure}[!ht]
\centerline{\includegraphics[width=0.99\textwidth]{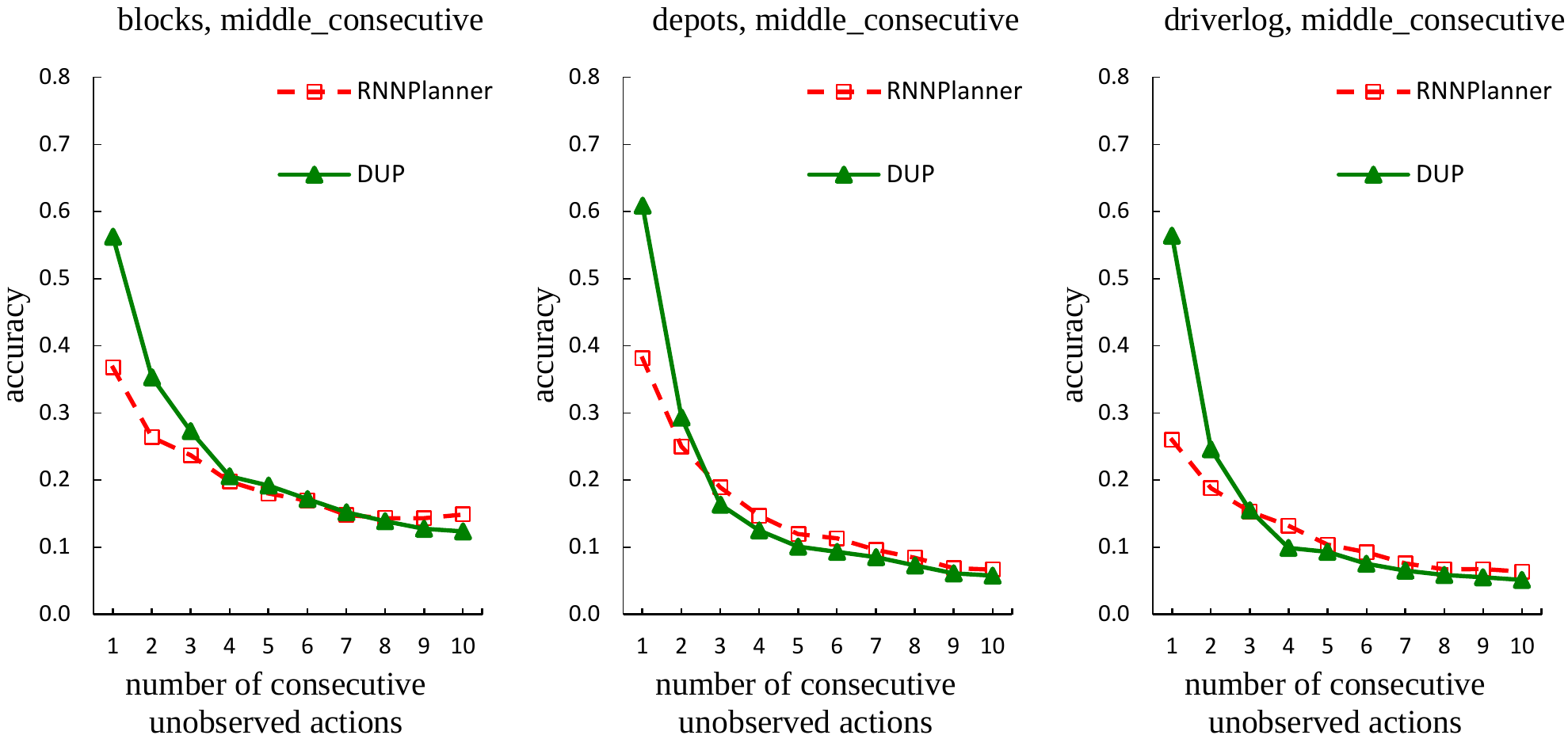}}
\caption{Accuracy with respect to missing actions in the middle}
\label{missing_middle}
\end{figure}

To see the performance of {\rnnplanner} and {\dup} in cases when actions are missing in the middle of the underlying plan to be discovered, we vary the number of consecutive missing actions from 1 to 10, to see the change of accuracies of both {\rnnplanner} and {\dup}. We set the window size to be 1 and the recommendation size to be 10. The results are shown in Figure \ref{missing_middle}. 

From Figure \ref{missing_middle}, we can see that the accuracies of both {\rnnplanner} and {\dup} generally become lower when the number of consecutive unobserved actions increasing. This is consistent with our intuition since the more actions are missing, the less information can be used to help infer the unobserved actions, which results in low accuracies. Comparing the curves of {\rnnplanner} and {\dup}, we can see that the accuracy of {\dup} is higher than {\rnnplanner} at the beginning. This is because {\dup} exploits information of both observed actions before and after missing actions to infer the missing actions, while {\rnnplanner} just exploits observed actions before missing actions. When the number of missing actions is larger than 3, the accuracies of {\dup} and {\rnnplanner} are both low (i.e., lower than 0.2). This is because the window size of {\dup} and {\rnnplanner} is set to be 1, which indicates we exploit one action before the missing actions and one action after the missing actions to estimate the missing actions. When the consecutive missing actions are more than 1, there may not be sufficient context information for inferring the missing actions, resulting in low accuracies. 

\subsubsection{Performance with missing actions at the end}
\begin{figure}[!ht]
\centerline{\includegraphics[width=0.99\textwidth]{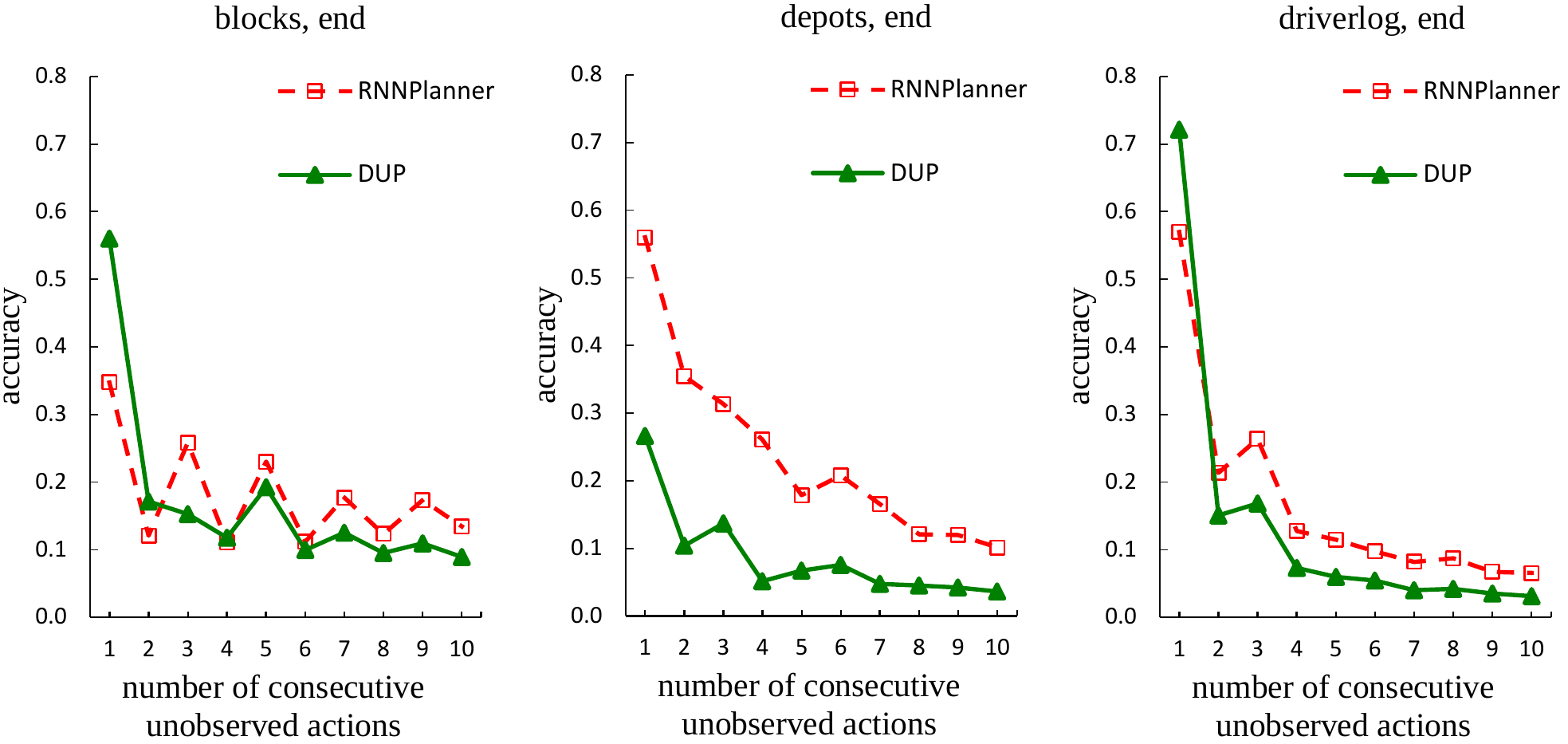}}
\caption{Accuracy with respect to missing actions in the end.}
\label{missing_end}
\end{figure}

We also would like to see the performance of {\rnnplanner} and {\dup} in discovering missing actions at the end, which is prevalent in application domains that aim at discovering/predicting future actions. Similar to previous experiments, we vary the number of consecutive unobserved or missing actions to see the change of accuracies of both {\rnnplanner} and {\dup}. We set the window size to be 1 and the recommendation size to be 10 as well. The experimental results are shown in Figure \ref{missing_end}.

From Figure \ref{missing_end}, we can observe that the accuracies of both {\rnnplanner} and {\dup} generally get decreasing when the number of consecutive missing actions increases. This is similar to previous experimental results. That is, the more actions are missing, the less information is available for estimating the missing actions, which results in lower accuracy. In addition, we can also see that {\rnnplanner} generally performs better than {\dup}, which indicates that the RNNs-based approach, i.e., {\rnnplanner}, can indeed better exploit observed actions to predict future missing actions, since RNNs are capable of flexibly leveraging long or short-term information to help predict missing actions. 

\subsubsection{Performance with respect to different recommendation size}
\begin{figure}[!ht]
\centerline{\includegraphics[width=1.0\textwidth]{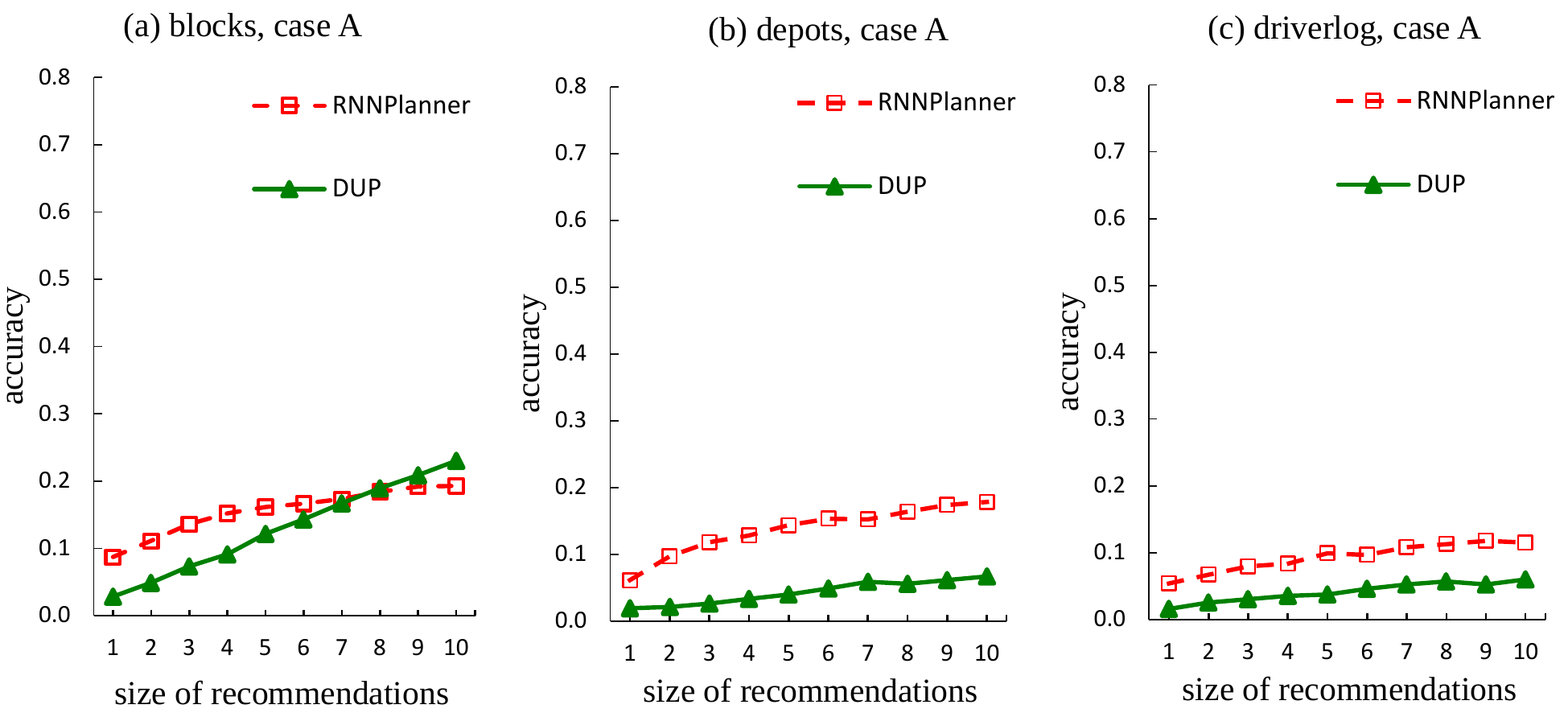}}
\caption{Case A: accuracy with respect to different size of recommendations}
\label{case_a_acc_rec}
\end{figure}

\begin{figure}[!ht]
\centerline{\includegraphics[width=1.0\textwidth]{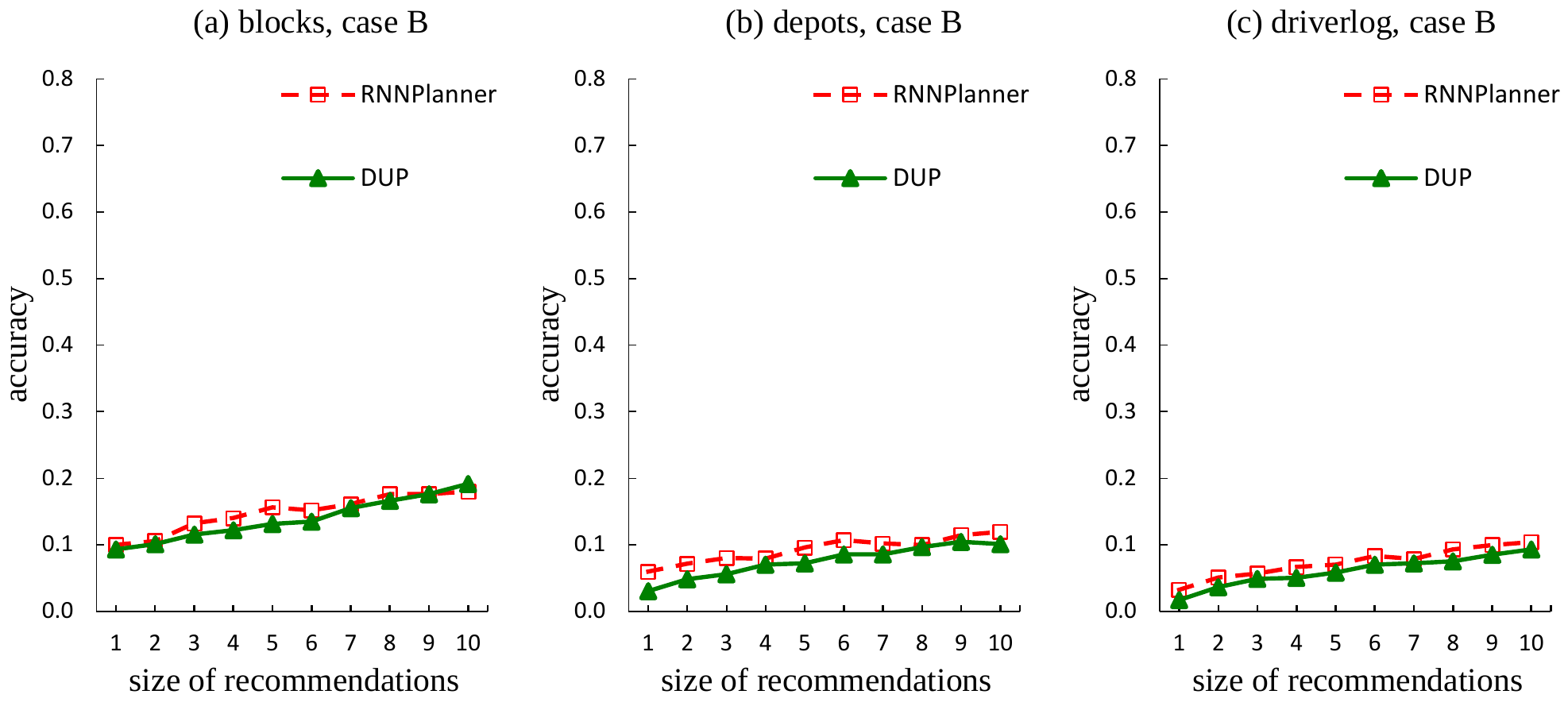}}
\caption{Case B: accuracy with respect to different size of recommendations}
\label{case_b_acc_rec}
\end{figure}

\begin{figure}[!ht]
\centerline{\includegraphics[width=1.0\textwidth]{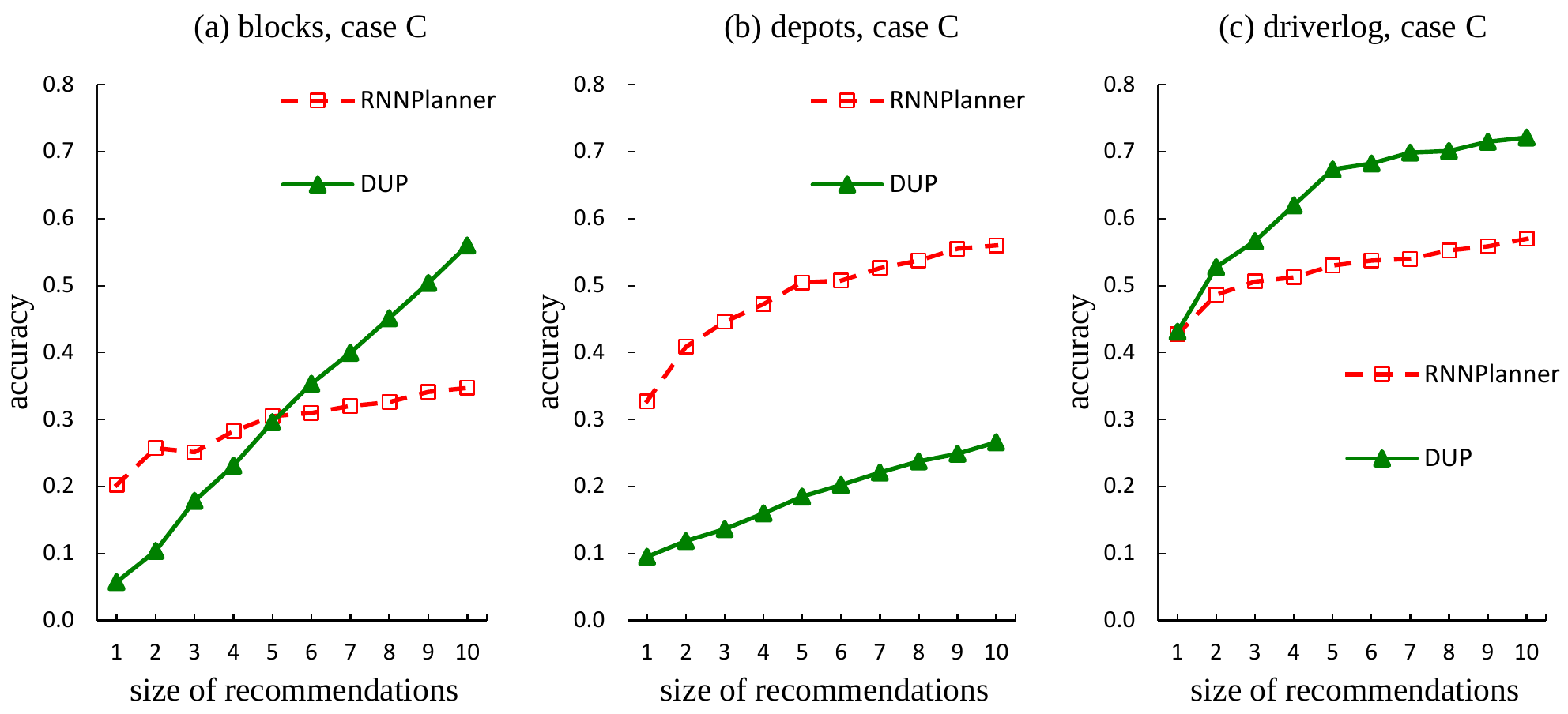}}
\caption{Case C: accuracy with respect to different size of recommendations}
\label{case_c_acc_rec}
\end{figure}

\begin{figure}[!ht]
\centerline{\includegraphics[width=1.0\textwidth]{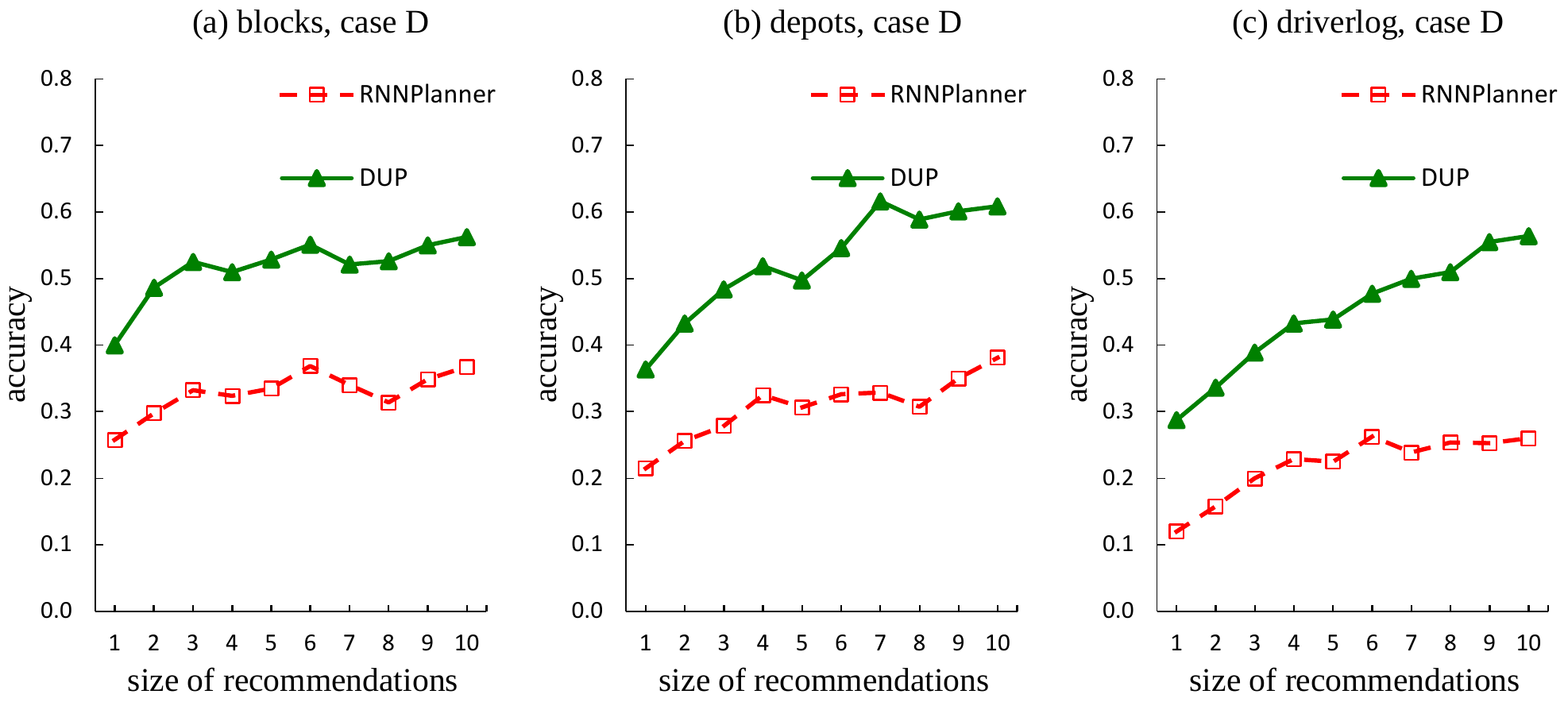}}
\caption{Case D: accuracy with respect to different size of recommendations}
\label{case_d_acc_rec}
\end{figure}

To see the change  with respect to different recommendation size, we vary the size of recommendation sets from 1 to 10 and calculate their corresponding accuracies. We test our approaches with four cases: A. there are five actions missing at the end; B. there are five actions missing in the middle; C. there is one action missing at the end; D. there is one action missing in the middle. The results are shown in Figures \ref{case_a_acc_rec}-\ref{case_d_acc_rec} corresponding to cases A-D, respectively.

\paragraph{Case A:} As shown in Figure \ref{case_a_acc_rec}, {\rnnplanner} performs better than {\dup} mostly, except for when the recommendation set size is larger/equal to eight in blocks domain. This is because  {\rnnplanner}, which contains LSTM cells, is able to actively remember or forget past observations (inputs) and computations (hidden states). For example, if in a set of sequences, a pattern $A^{**}$ follows $A^*$ after three words (i.e., $..., A^*, w_i, w_{i+1}, w_{i+2}, A^{**}, ...$), where $w_i, w_{i+1}, w_{i+2}$ could be any word from the vocabulary except for $A^{*}$ and $A^{**}$. And if the window size of {\dup} is smaller than three, then {\dup} is not able to utilize this pattern to predict $A^{**}$ mainly based on $A^*$. When predicting the $A^{**}$, the {\dup} with context size one, works by searching for the most similar word to the $w_{i+2}$. One would yet argue that we can set a larger window size for {\dup}. Larger window size does not necessarily lead to higher accuracy, since using a larger window size also add more noise in the training of {\dup}. Remember that the word2vec model treats equally all possible word pair samples within its context window.

In addition, observing the accuracies (in terms of the size of recommendation $S_x$) of all three domains , we can see only in the blocks domain that {\dup} outperforms {\rnnplanner}, when $S_x$ is larger than eight. Also in the \emph{blocks} domain, {\dup} has the best performance, comparing to how {\dup} functions in other two domains. This is because plans from the \emph{blocks} domain has an overall higher ratio of \#word to \#vocabulary, which increases the possibility that the word pattern outside a context window, would reappear inside the window, and consequently help {\dup} recognize actions in the missing positions. Coming back to the example when we have a plan like $..., A^*, w_i, w_{i+1}, w_{i+2}, A^{**}, ...$, in \emph{blocks} domain, it's more possible the word $A^*$ happens again in one of $w_i$, $w_{i+1}$, and $w_{i+2}$.

\paragraph{Case B:} What we can observe here from Figure \ref{case_b_acc_rec}, is similar to our observations in {\caseA}. {\rnnplanner} generally performs better than {\dup}, except for when the size of recommendation $S_x$ is larger or equal to nine in \emph{blocks} domain. It could also be observed that both {\rnnplanner} and {\dup} have the best accuracy performance in the \emph{blocks} domain. 

And by comparing the Figure \ref{case_b_acc_rec} in {\caseB} (five removed actions in the middle) with Figure \ref{case_a_acc_rec} in {\caseA} (five removed actions at the end), we can see that, the accuracy difference between {\dup} and {\rnnplanner} at each size of recommendation along the x-axis, is smaller in {\caseB}. This is because, {\rnnplanner} only leverages the observed actions before a missing position, whereas {\dup} has the advantage of additionally using the observation after a missing position.

\paragraph{Case C:} From Figure \ref{case_c_acc_rec}, we can see that both {\rnnplanner} and {\dup} could outperform each other in certain domains and recommendation set sizes ($S_x$). In \emph{blocks} domain, {\dup} is better when $S_x$ is larger than five. In \emph{depots} domain, {\rnnplanner} is overwhelmingly better than {\dup}. In \emph{driverlog} domain, {\dup} performs overall better except that, when there is only one recommendation, {\dup} is as good as {\rnnplanner}. To explain this observation, if the number of consecutively removed action is less or equal to context window size (e.g., window size is one, and number of missing actions is one, in our {\caseC}), then the fixed, and short context window of {\dup}, is competitive enough.

\paragraph{Case D:} From the results in Figure \ref{case_d_acc_rec}, we can see that {\dup} functions better than {\rnnplanner} over all three domains, whereas {\dup} is worse in {\caseA} and {\caseB}, and could occasionally be better than {\rnnplanner} in {\caseC}. It makes sense in that, on the one hand, within the fixed and short context window, if there is very less positions with removed actions, {\dup} would have an improved performance. On the other hand, {\rnnplanner} is not able to leverage the information from both sides of a position with a missing action. Therefore, in {\caseD}, {\dup} gains the benefit from both assumptions that there is only one missing action, and the position of that action is randomly chosen in the middle of a plan. 

\section{Related work}
\begin{figure*}[!ht]
\centerline{\includegraphics[width=1.0\textwidth]{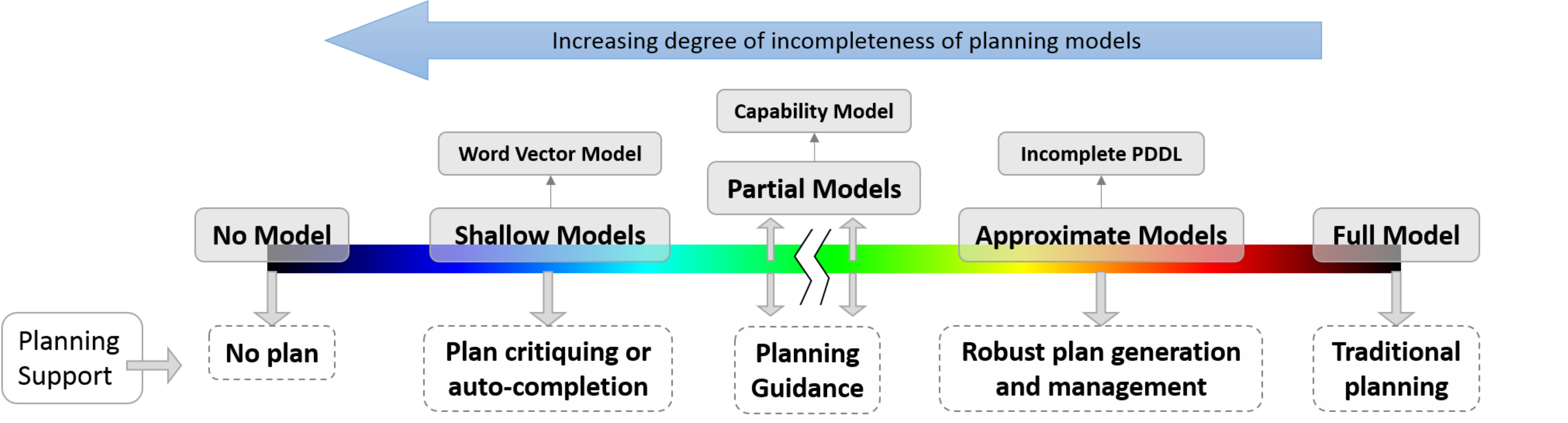}}
\caption{Schematic view of incomplete models and their relationships in the spectrum of incompleteness}
\label{spectrum}
\end{figure*}

Our work is related to planning with incomplete domain models (or model-lite planning \cite{conf/aaai/rao07,DBLP:conf/aaai/ZhuoNK13}).
Figure \ref{spectrum} shows the schematic view of incomplete models and their relationships in the spectrum of incompleteness. In a full model, we know exactly the dynamics of the model (i.e., state transitions). Approximate models are the closest to full
models and their representations are similar
except that there can be incomplete knowledge of action descriptions. To enable approximate planners to
perform more (e.g., providing robust plans),
planners are assumed to have access
to additional knowledge circumscribing the
incompleteness \cite{tuan-robust}. Partial models are one level further down the line in terms of the degree of incompleteness. While approximate models can encode incompleteness in the precondition/effect descriptions of the individual actions, partial models can completely abstract portions of a plan without providing details for them. In such cases, even though providing complete plans is infeasible, partial models can provide ``planning guidance'' for agents \cite{DBLP:conf/atal/ZhangSK15}. Shallow models are essentially just a step above having no planning model. They provide interesting contrasts to the standard precondition and effect based action models used in automated planning community. Our work in this paper belongs to the class of shallow models. In developing shallow models, we are interested in planning technology that helps humans develop plans, even in the absence of any structured models or plan traces. In such cases, the best that we can hope for is to learn local structures of the planning model to provide planning support, similar to providing spell-check in writing. While some work in web-service composition (c.f. \cite{DBLP:conf/vldb/DongHMNZ04}) did focus on this type of planning support, they were hobbled by being limited to simple input/output type comparison. In contrast, we expect shallow models to be useful in ``critiquing'' the plans being generated by the humans (e.g. detecting that an action introduced by the human is not consistent with the model), and ``explaining/justifying'' the suggestions generated by humans.  

Our work is also related to plan recognition. Kautz and Allen proposed an approach to recognizing plans based on parsing observed actions as sequences of subactions and essentially model this knowledge as a context-free rule in an ``action grammar'' \cite{cof/aaai/kautz86}. All actions, plans are uniformly referred to as goals, and a recognizer's knowledge is represented by a set of first-order statements called event hierarchy encoded in first-order logic, which defines abstraction, decomposition and functional relationships between types of events. Lesh and Etzioni further presented methods in scaling up activity recognition to scale up his work computationally \cite{DBLP:conf/ijcai/LeshE95}. They automatically constructed plan-library from domain primitives, which was different from \cite{cof/aaai/kautz86} where the plan library was explicitly represented. In these approaches, the problem of combinatorial explosion of plan execution models impedes its application to real-world domains. Kabanza and Filion \cite{DBLP:conf/ijcai/KabanzaFBI13} proposed an anytime plan recognition algorithm to reduce the number of generated plan execution models based on weighted model counting. These approaches are, however, difficult to represent uncertainty. They offer no mechanism for preferring one consistent approach to another and incapable of deciding whether one particular plan is more likely than another, as long as both of them can be consistent enough to explain the actions observed.\ignore{ Bui et al. \cite{cof/ijcai/Bui03,journal/aij/Geib09} presented approaches to probabilistic plan recognition problems. } Although we exploit a library of plans in our {\tt DUP} and {\rnnplanner} approaches, we aim to learning shallow models and utilize the shallow models to recognize plans that are not necessarily in the plan library, which is different from previous approaches that assume the plans to be recognized are from the plan library.

Instead of using a library of plans, Ramirez and Geffner \cite{cof/ijcai/Ramirez09} proposed an approach to solving the plan recognition problem using slightly modified planning algorithms, assuming the action models were given as input (note that action models can be created by experts or learnt by previous systems \cite{journal/aij/Yang07,journal/aij/zhuo10}). Except previous work \cite{cof/aaai/kautz86,cof/ijcai/Bui03,journal/aij/Geib09,cof/ijcai/Ramirez09} on the plan recognition problem presented in the introduction section, \ignore{Note that action models can be created by experts or learnt by previous systems, such as {\tt ARMS} \cite{journal/aij/Yang07} and {\tt LAMMAS} \cite{DBLP:conf/atal/ZhuoMY11}.} Saria and Mahadevan presented a hierarchical multi-agent markov processes as a framework for hierarchical probabilistic plan recognition in cooperative multi-agent systems \cite{conf/ICAPS/Saria04}. \ignore{Singla and Mooney proposed an approach to abductive reasoning using a first-order probabilistic logic to recognize plans \cite{conf/aaai/Singla11}. }Amir and Gal addressed a plan recognition approach to recognizing student behaviors using virtual science laboratories \cite{cof/ijcai/Amir11}. Ramirez and Geffner exploited off-the-shelf classical planners to recognize probabilistic plans \cite{cof/aaai/Ramirez10}. Different from those approaches, we do not require any domain model knowledge provided as input. Instead, we automatically learn shallow domain models from previous plan cases for recognizing unknown plans that may not be identical to previous cases. 

\section{Conclusion and Discussion}

In this paper we present two novel plan recognition approaches, {\dup} and {\rnnplanner},
based on vector representation of actions. For {\dup}, we first learn the vector
representations of actions from plan libraries using the Skip-gram
model which has been demonstrated to be effective. We then discover
unobserved actions with the vector representations by repeatedly
sampling actions and optimizing the probability of potential plans to
be recognized. For  {\rnnplanner}, we let the neural network itself to learn the word embedding, which would then be utilized by higher LSTM layers. We also empirically exhibit the effectiveness of our
approaches. 

While we focused on a one-shot recognition task in this paper, in
practice, human-in-the-loop planning will consist of multiple
iterations, with DUP recognizing the plan and suggesting action
addition alternatives; the human making a selection and revising the
plan. The aim is to provide a form of flexible plan completion tool,
akin to auto-completers for search engine queries. To do this
efficiently, we need to make the DUP recognition algorithm ``incremental.''

The word-vector based domain model we developed in this paper provides
interesting contrasts to the standard precondition and effect based
action models used in automated planning community. One of our future
aims is to provide a more systematic comparison of the tradeoffs
offered by these models.  Although we have
focused on the ``plan recognition'' aspects of this model until now,
and assumed that ``planning support'' will be limited to suggesting
potential actions to the humans. In future, we will also consider
``critiquing'' the plans being generated by the humans (e.g. detecting
that an action introduced by the human is not consistent with the
model learned by DUP), and ``explaining/justifying'' the suggestions
generated by humans. Here, we cannot expect causal explanations of the
sorts that can be generated with the help of complete action models
(e.g. \cite{petrie}), and will have to develop justifications
analogous to those used in recommendation systems. 

Another potential application for this type of distributed action representations proposed in this paper is social media analysis. In particular, work such as \cite{kiciman2015} shows that identification of action-outcome relationships can significantly improve the analysis of social media threads. The challenge of course is that such action-outcome models have to be learned from raw and noisy social media text containing mere fragments of plans. We believe that action vector models of the type we proposed in this paper provide a promising way of handling this challenge. 

\begin{acknowledgements}
Zhuo thanks the support of the National Key Research and Development Program of China (2016YFB0201900), National Natural Science Foundation of China (U1611262), Guangdong Natural Science Funds for Distinguished Young Scholar (2017A030306028), Pearl River Science and Technology New Star of Guangzhou, and Guangdong Province Key Laboratory of Big Data Analysis and Processing for the support of this research. Kambhampati's research is supported  in part by the ARO grant
W911NF-13-1-0023, the ONR grants N00014-13-1-0176, N00014-09-1-0017
and N00014-07-1-1049, and the NSF grant IIS201330813.
\end{acknowledgements}

\bibliographystyle{spmpsci}      
\bibliography{JAAMAS}   

\end{document}